%% file: main.tex
\documentclass[lettersize,journal]{IEEEtran}
\usepackage{amsmath,amsfonts}
\usepackage{listings}
\usepackage{algorithmic}
\usepackage{bm}
\usepackage[ruled, vlined, linesnumbered]{algorithm2e}
\usepackage{array}
\usepackage{textcomp}
\usepackage{stfloats}
\usepackage{url}
\usepackage{verbatim}
\usepackage{graphicx}
\usepackage{cite}
\usepackage{multirow}
\usepackage{makecell}
\usepackage{booktabs}
\usepackage{color}
\usepackage{subcaption}
\usepackage{pifont}
\usepackage[flushleft]{threeparttable}  % for adding footnotes to table
\usepackage[table,dvipsnames]{xcolor}
\usepackage{wrapfig} % for wrap text around a table
\usepackage{xcolor}
\usepackage{multicol}
\usepackage{pifont}
\usepackage{tcolorbox}
\usepackage{tabularx}
\usepackage[normalem]{ulem}

\newif\iffinal

\lstdefinestyle{pythonstyle}{
  columns=fullflexible,
  breaklines=true,
  captionpos=b,
  xleftmargin=1em,
  backgroundcolor=\color{white},
  commentstyle=\color{green},
  keywordstyle=\color{magenta},
  numberstyle=\tiny\color{gray},
  stringstyle=\color{purple},
  basicstyle=\ttfamily\footnotesize,
  breakatwhitespace=false,
  breaklines=true,
  keepspaces=true,
  numbers=left,
  numbersep=5pt,
  showspaces=false,
  showstringspaces=false,
  showtabs=false,
  tabsize=2
}

\begin{document}

\title{Exploring the True Potential: Evaluating the Black-box Optimization Capability of Large Language Models}
%Effective or Trendy Abuse: Evaluating the True Optimization Capability of LLM

\author{
    Beichen Huang,
    Xingyu Wu,
    Yu Zhou,
    Jibin Wu,
    Liang Feng,~\IEEEmembership{Senior Member,~IEEE}
    Ran Cheng,~\IEEEmembership{Senior Member,~IEEE},
    Kay Chen Tan,~\IEEEmembership{Fellow,~IEEE}
    \thanks{Beichen Huang, Xingyu Wu, Yu Zhou, Jibin Wu, and Kay Chen Tan are with the Department of Computing, The Hong Kong Polytechnic University, Hong Kong SAR 999077, China. (E-mails: beichen.huang@connect.polyu.hk, xingy.wu@polyu.edu.hk, zy-yu.zhou@connect.polyu.hk, jibin.wu@polyu.edu.hk, kaychen.tan@polyu.edu.hk)}
    \thanks{Ran Cheng is with the Department of Computer Science and Engineering, Southern University of Science and Technology, Shenzhen 518055, China. (E-mail: ranchengcn@gmail.com)}
    \thanks{Liang Feng is with the College of Computer Science, Chongqing University, Chongqing 400044, China. (E-mail: liangf@cqu.edu.cn)}
}

% \email{beichen.huang@connect.polyu.hk}
% \affiliation{%
%   \institution{Hong Kong Polytechnic University}
%   \city{Hong Kong}
%   \country{Hong Kong SAR}
%   \postcode{000000}
% }

\markboth{}
{Shell \MakeLowercase{\textit{et al.}}}

\maketitle

\input{0-abstract}

\input{1-introduction}
\input{2-related}

\input{3-experiment}
\input{4-conclusion}

\bibliographystyle{IEEEtran}
\bibliography{egbib}
\input{5-appendix}

\end{document}

%% file: 0-abstract.tex
\begin{abstract}
Large language models (LLMs) have demonstrated exceptional performance not only in natural language processing tasks but also in a great variety of non-linguistic domains. 
In diverse optimization scenarios, there is also a rising trend of applying LLMs.  
However, whether the application of LLMs in the black-box optimization problems is genuinely beneficial remains unexplored. 
This paper endeavors to offer deep insights into the potential of LLMs in optimization through a comprehensive investigation, which covers both discrete and continuous optimization problems to assess the efficacy and distinctive characteristics that LLMs bring to this field.
Our findings reveal both the limitations and advantages of LLMs in optimization.
Specifically, on the one hand, despite the significant power consumed for running the models, LLMs exhibit subpar performance
in pure numerical tasks, primarily due to a mismatch between the problem domain and their processing capabilities;
on the other hand, although LLMs may not be ideal for traditional numerical optimization, their potential in broader optimization contexts remains promising, where LLMs exhibit the ability to solve problems in non-numerical domains and can leverage heuristics from the prompt to enhance their performance.
To the best of our knowledge, this work presents the first systematic evaluation of LLMs for numerical optimization. Our findings pave the way for a deeper understanding of LLMs' role in optimization and guide future application of LLMs in a wide range of scenarios.
\end{abstract}

%% file: 1-introduction.tex
\section{Introduction}
Large language models (LLMs) have rapidly gained tremendous popularity since their inception, with remarkable performance in natural language processing (NLP) tasks~\cite{min_recent_2021,zhao_survey_2023,minaee_large_2024,blair-stanek_can_2023,tang_does_2023} as well as lots of other domains beyond NLP~\cite{beltagy_scibert_2019,xie_darwin_2023,thirunavukarasu_large_2023,boiko_autonomous_2023}.  
In the field of optimization, LLMs are also applied incrementally~\cite{yang_large_2023,guo_towards_2023,meyerson_language_2023,liu_large_single_2023,lange_large_2024,huang_how_2024,liu_large_multi_2023,nasir_llmatic_2023,wang_large_2024,brahmachary_large_2024,wu2024evolutionary}.
However, with limited scope and depth as well as a small number of explored tasks,
it is still unclear whether the application of LLMs to non-linguistic optimization problems is driven by the current trend of massively applying LLMs or if they genuinely offer distinct advantages in solving these problems. 
A thorough investigation is necessary to establish the validity and reliability of LLMs in this context.

In this work, we conduct a comprehensive evaluation to reveal the true potential of LLMs in diverse optimization tasks. 
Our investigation focuses on the performance of LLMs across different classic black-box optimization problems, encompassing both discrete and continuous optimization domains, to uncover the unique characteristics that LLMs manifest during the optimization process as much as possible.
Furthermore, we seek to understand the mechanisms of using LLMs for solving optimization problems by examining their fundamental properties in comparison to traditional algorithms and also those beyond.
To the best of our knowledge, this work is the first systematic evaluation of LLMs' capabilities in solving {numerical black-box optimization} problems.

\begin{figure*}
    \centering
    \includegraphics[width=\linewidth]{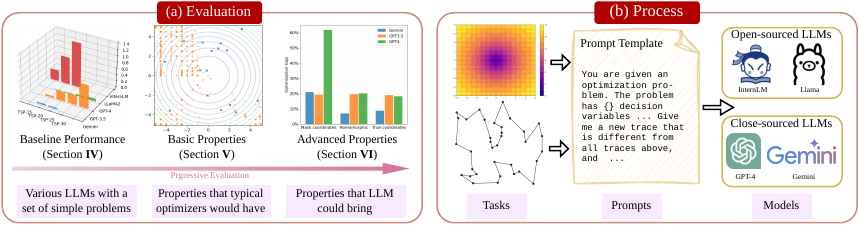}
    \caption{An illustration of our evaluation on applying popular LLMs for black-box optimization.
    %It encompasses multiple LLMs and tasks, progressing from basic properties to advanced properties. 
    (a) Overall evaluation process. First, we will assess the baseline performance of different models through a series of simple tasks. These baseline experiments will help identify a subset of top-performing models, which will then be utilized to evaluate the essential properties of optimizers on LLMs in detail. We will begin by examining basic properties, which are typical of most optimizers, and subsequently move on to advanced properties that only LLMs may possess. (b) Process of evaluating each property. First, we will design a task that can reflect the property in question. Next, we will create a prompt template tailored to this task. Finally, we will employ multiple models to conduct the optimization process, thereby assessing their performance regarding the evaluated property.}
    \label{fig:process}
\end{figure*}

Our evaluation features a progressive evaluation process. As shown in Fig.~\ref{fig:process} (a), we first test various LLMs with simple optimization problems to check their baseline performance as optimizers, based on which we identify the top-performing LLMs to be analyzed further. 
We then examine the properties of these models in solving optimization problems, first focusing on those basic properties that are typical to traditional optimizers, and then proceeding to those that are unique to LLM-based optimizers.
In each evaluation, as shown in  Fig.~\ref{fig:process} (b), we follow a classic approach. Namely, we design a task that can reflect each property and craft a corresponding prompt template, which is then used to query the LLMs and execute the optimization task.

More specifically, we speculate that an LLM should demonstrate certain essential properties when used to solve numerical optimization tasks, including but not limited to:
\begin{itemize}
    \item The ability to understand numerical values (Section~\ref{subsec:number}); 
    \item The ability to handle multidimensional vector data
    (Section~\ref{subsec:scale}), as well as {the scalability on problem dimensions} (Section~\ref{sec:baseline} and Section~\ref{subsec:scale});
    \item The adaptability to diverse problem scenarios (Section~\ref{subsec:shift}); 
    \item The balance between exploration and exploitation (Section~\ref{subsec:balance}), as well as a well-formed optimization process (Section~\ref{subsec:balance}).
\end{itemize}

The first property is inherent to traditional optimizers since their implementation utilizes integer or floating point arithmetic, thus guaranteeing its validity. The remaining properties are fundamental to traditional optimizers, as they were designed with these characteristics in mind~\cite{review_blackbox_part1,review_blackbox_part2,zhan_survey_2022}.
%\textcolor{red}{The distinction between the words `inherent' and `basic' seems to be small. Both seem to be essential properties to me. To make sure you have communicated your idea clearly, please consider to replace with other words. } In this work, we examine them on a variety of popular LLMs with detailed experiments.  

{In addition to the basic properties,  we are also interested in what LLMs could achieve with their unique capabilities beyond the reach of traditional optimization algorithms. 
Conventional algorithms rely on expert knowledge, which is hardcoded into their algorithm implementation as heuristics, to help address specific optimization problems.  
As a result, they tend to be specialized and limited in their scope. 
In comparison, LLM-based algorithms possess knowledge from diverse domains, and they make decisions based on prompt engineering.
Thus, LLM-based optimizers are less restricted by fixed logics, and can potentially generate heuristics on their own to help problem-solving.
In this work, we test whether LLMs can enhance their optimization capabilities by leveraging available information from the prompt, {without being directly guided by any expert knowledge, i.e., without relying on human-written code or explicit hints in the prompt to guide their behavior.} 
We call these properties ``the advanced properties'' (Section~\ref{sec:advanced})}.

To ensure a comprehensive investigation of the multifaceted abilities of LLMs, our evaluation includes a wide coverage of optimization tasks.
In particular, our evaluation spans both discrete and continuous types of optimization problems, each comprising multiple tasks.
By examining LLMs' performance in diverse problem domains, we seek to provide a full perspective on the adaptability and effectiveness of LLMs as optimization tools.
Besides, we also seek to perform behavioral analysis for LLM-based optimizers. 
Beyond merely analyzing optimization results, we take a deep dive into the behavioral aspects of LLMs during optimization, analyzing the patterns in LLMs' actions. This analysis offers valuable insights into the strengths and weaknesses of LLMs, shedding light on their functioning mechanisms. Such an exploration goes beyond conventional assessments, providing a richer understanding of how LLMs operate in black-box optimization contexts.

Building upon the advantageous evaluation featured with a progressive evaluation process, wide coverage of optimization types as well as behavioral analysis, we acquire some insightful conclusions about the capabilities of LLMs in tackling black-box optimization problems, which will, in turn, inform and guide our future research directions. The key takeaways from our conclusions are:

\begin{itemize}
    \item \textbf{LLMs are still less competent for tackling numerical optimization problems.}
    LLMs prove less suitable for direct engagement in pure numerical optimization tasks, given their reliance on string representation. 
    They lack some essential properties of the effective optimizers, demonstrating limited capabilities in utilizing floating-point numbers as shown in Section~\ref{subsec:number}, handling multidimensional vector data as shown in Section~\ref{subsec:scale}, scaling as examined in Section~\ref{sec:baseline} and Section~\ref{subsec:scale}, adapting to shift variant problems as revealed in Section~\ref{subsec:shift}, balancing between exploration and exploitation as shown in Section~\ref{subsec:balance}, and forming adequate generation patterns as shown in Section~\ref{subsec:balance}. 
    The absence of these properties emphasizes the need for caution when applying LLMs to solve optimization problems.

    \item \textbf{LLMs offer boosted performance in some specific scenarios.}
    Despite their limitations in numerical optimization, LLMs exhibit distinct advantages over traditional algorithms in specific scenarios. In contrast to hand-crafted algorithms, LLMs eliminate the need for human intervention to model problems in mathematical formats or solve them manually, as shown in both Sections~\ref{sec:basic} and \ref{sec:advanced} where no explicit instructions about the optimization problem or the solving steps are provided. Furthermore, in Section~\ref{subsec:coord}, LLMs demonstrate the ability to extract additional information from the problem description itself, suggesting that LLMs can potentially generate heuristics naturally tailored to specific problems, despite defective generation patterns as shown in Section~\ref{subsec:coord} and \ref{subsec:cityname}.

    \item \textbf{LLMs have a promising future in the optimization field.} In Section~\ref{sec:conclusion}, we embark on a thoughtful exploration of the future trajectory of LLMs in the optimization field. Our analysis, incorporating their weaknesses, potential advancements, and challenges, is discussed in Section~\ref{sec:basic} and Section~\ref{sec:advanced}. While acknowledging the significant potential of LLMs in optimization, it is crucial to note that their strengths may not lie in traditional pure numerical optimization. This limitation arises from the mismatch between traditional problem formulation and LLM processing capabilities. To address this issue, we propose integrating external tools into LLMs specifically designed for computing tasks, rather than having LLMs directly handle numerical data. On the other hand, we must point out that LLMs' performance in these areas does not negate their potential future role in the field of optimization. It is essential to not limit our vision to numerical benchmark problems, and adopt a broader perspective on optimization, envisioning scenarios where LLMs can indeed excel and contribute meaningfully. For example, LLM could be important for optimization problems in the text domain, e.g., prompt engineering and code generation. 
\end{itemize}

The remainder of this paper is structured as follows. We first briefly review the related works in Section~\ref{sec:related}. This is followed by the descriptions of our experiment settings in Section~\ref{sec:settings}. In Section~\ref{sec:Investigation}, we investigate and analyze LLMs in black-box optimization tasks. Finally, we conclude the paper in Section~\ref{sec:conclusion} by summarizing our findings and discussing future prospects of LLMs in optimization.

%% file: 2-related.tex
\section{Related Work \label{sec:related}}

Our research aims to assess the effectiveness of employing LLMs in the realm of black-box optimization. In this section, we first provide a brief review of previous studies on LLMs, and then focus on their applications to numerical black-box optimization problems.

\subsection{Large Language Models}
LLMs are powerful natural language processing tools that have revolutionized the field with remarkable abilities to understand and generate human-like text. 
These models are often characterized by massive scales, comprising billions or trillions of parameters, and have demonstrated state-of-the-art performance in various language-related tasks.
Among them, some LLMs (usually named the chat models) are tuned to handle natural chat-like interaction, where the user describes the task in natural language, and the model gives the answer as a response.
Currently, several influential LLMs have been developed and quickly gained explosive popularity.
To name a few: 

\begin{itemize}
    \item \textbf{BERT}~\cite{bert_2018}: Bidirectional Encoder Representations from Transformers (BERT) is a powerful language model created by Google in 2018. It is known for its significant effectiveness in solving NLP tasks. BERT is pre-trained with a large amount of unlabeled texts and able to fit a wide range of downstream tasks, with just an additional output layer.
    
    \item \textbf{T5}~\cite{t5_2023}: Text-To-Text Transfer Transformer (T5) is a versatile language model introduced by Google Research. 
    T5 adopts a unified framework where all NLP tasks are framed as text-to-text problems, demonstrating flexibility and effectiveness across various language-related tasks.

    \item \textbf{GPT-3}: Generative Pre-training Transformer 3 (GPT-3) is developed by OpenAI, standing as one of the largest and most powerful language models, boasting 175 billion parameters.
    In 2022, GPT-3 was further optimized to GPT-3.5, and at the end of 2022, ChatGPT\footnote{chat.openai.com} services were launched backed by GPT-3.5.

    \item \textbf{GPT-4}~\cite{openai_gpt-4_2024}: Building on the success of GPT-3, OpenAI's GPT-4 continues the trend of scaling language models.
    Although specific details may vary, GPT-4 is anticipated to push the boundaries of language understanding and generation, addressing challenges and limitations observed in its predecessor.
    GPT-4 became publicly accessible in March 2023 through a paid service.

    \item \textbf{LLaMA}~\cite{touvron_llama_2023}: Large Language Model Meta AI (LLaMA) is a family of powerful AI models trained on publicly available datasets. These open-source models achieve state-of-the-art performance in various natural language processing tasks. Moreover, LLAMA models are open-sourced, with model and pre-trained parameters available to the public. In July 2023, LLaMA2~\cite{touvron_llama2_2023} was released as the next-generation model in this family.

    \item \textbf{Alpaca}~\cite{wang_self-instruct_2023}: Alpaca is a fine-tuned version of Meta’s LLaMA 7B model from Standford using the self-instruct method. Alpaca is an instruction-following model and is open-sourced for all research purposes.

    \item \textbf{Gemini}~\cite{gemini_team_gemini_2023}: Unveiled in 2023 by Google DeepMind, Gemini is a family of powerful LLMs that is native multimodal jointly across text, image, audio, and video. The family of models demonstrates impressive capabilities across a broad spectrum of tasks.
    Gemini is available for public access through Google's online services.

    \item \textbf{InternLM}: InternLM\footnote{github.com/InternLM/InternLM} is a language model with a substantial scale of 20 billion parameters, developed by Shanghai Artificial Intelligence Laboratory in collaboration with SenseTime Technology, the Chinese University of Hong Kong, and Fudan University. The model is publicly available and its pre-trained model is trained on an extensive dataset comprising over 2.3 trillion tokens, encompassing high-quality English, Chinese, and code data.
\end{itemize}

\begin{table}[h]
  \centering
  \caption{A summary of popular LLMs}
  \begin{tabular}{c c c c}
    \toprule
    \textbf{Model} & \textbf{Developer(s)} & \textbf{Year} & \textbf{Open-sourced}
    \\
    \midrule
    BERT & Google & 2018 & Yes\\
    T5 & Google & 2019 & Yes\\
    GPT-3 & OpenAI & 2020 & No\\
    GPT-4 & OpenAI & 2023 & No\\
    Alpaca & Research Team & 2022 & Yes\\
    LLaMA2 & Meta & 2023 & Yes\\
    Gemini & Google & 2023 & No \\
    InternLM & Research Team & 2023 & Yes\\
    \bottomrule
  \end{tabular}

  \label{tab:llm_summary}
\end{table}

We summarize the most popular LLMs in Table~\ref{tab:llm_summary}.
Note that LLMs are experiencing continuous innovations as time goes on. 
As the researchers consistently devote efforts to exploring new architectures, training strategies, or other enhancements, the capabilities of LLMs will be further lifted for language understanding, generation, etc.

\subsection{LLMs for Numerical Black-Box Optimization}

Black-box optimization, a longstanding and challenging field of optimization, involves optimizing an objective function whose underlying structure is entirely unknown. Particularly, numerical black-box optimization problems, where inputs and outputs are numerical data types, have garnered significant attention from researchers. Over the years, lots of algorithms have been proposed, which are typically designed by experts with extensive knowledge of the specific problem domain. However, a recent breakthrough in LLMs has led to a surprising discovery: these language models can be directly utilized to solve numerical black-box optimization problems. This innovative approach marks a significant and unprecedented trend in the field, as it enables the use of a general-purpose language model to optimize complex systems, without requiring expert knowledge of the underlying objective function.

Here we give a review of the works utilizing LLMs for numerical black-box optimization.
These works can be divided into two categories, single-objective and multi-objective optimization. 
{In single-objective settings, the objective is a single scalar value that can be compared and ordered, enabling a direct evaluation of the solution quality by simply comparing the objective value. Comparatively, multi-objective settings involve objectives with multiple values, which are only partially ordered. In this context, a solution is considered dominant if all of its objective values surpass those of another solution. Conversely, if some objective values are better while others are worse, the two solutions are deemed as in a non-dominated relation, where neither solution is truly superior to the other.}

\textbf{Single-Objective Optimization:}
Yang et al.~\cite{yang_large_2023} introduced the innovative concept of the Large Language Model as an Optimizer (OPRO). In OPRO, the LLM receives a meta-prompt as input and generates new solutions as output. The meta-prompt is updated throughout the optimization process to include the best solutions and their corresponding scores. As optimization progresses, the LLM receives increasingly up-to-date prompts, enabling it to generate better solutions and drive the process forward.
Building on this work, several new approaches have been proposed to integrate LLMs into existing optimization algorithm frameworks.
Liu et al.~\cite{liu_large_single_2023} and Meyerson et al.~\cite{meyerson_language_2023} integrated the LLM as operators within an evolutionary algorithm framework. Liu et al. focused on generating high-quality solutions through rigorous selection, crossover, and mutation processes, while Meyerson et al. utilized LLMs' text processing abilities to allow evolutionary algorithms to optimize text data, such as mathematical expressions, English sentences, and code.
Lange et al.~\cite{lange_large_2024} integrated the LLM into the evolution strategies~\cite{hansen2015evolution} framework and proposed Large Language Models As Evolution Strategies (EvoLLM). In this approach, the LLM serves as a recombination operator, processing a list of sorted solutions to propose a new mean value for the next iteration.
Several methods have been proposed to enhance the optimization abilities of LLMs.
Guo et al.~\cite{guo_towards_2023} extended the application of LLMs beyond black-box optimization to include gradient descent, hill climbing, and grid search settings.
Huang et al.~\cite{huang_how_2024} incorporated multimodel data into the optimization process, i.e. both text and images to the input of LLMs with multimodel capabilities. They conducted a comprehensive empirical study on the classic Capacitated Vehicle Routing Problem (CVRP)~\cite{toth2014vehicle}, and suggested that incorporating multimodal data can improve the performance of CVRP optimization.
Lange et al.~\cite{lange_large_2024} proposed a novel strategy to circumvent the LLMs' context length limitation. Normally, an LLM query in EvoLLM requires submitting all solutions as text in a single query. However, this strategy allows splitting the solutions along their problem dimensions, performing a batch of LLM queries, and subsequently aggregating the results to obtain the full answer.

\textbf{Multi-Objective Optimization:}
LLM can serve as a multi-objective solver when integrated into existing multi-objective algorithm frameworks.
Liu et al.~\cite{liu_large_multi_2023} integrated the LLM as an operator within a decomposition-based multi-objective optimization framework, while Bradley et al.~\cite{nasir_llmatic_2023} integrated LLM Quality-Diversity (QD) search~\cite{pugh2016quality} and proposed QD through AI Feedback (QDAIF) algorithm. Both works have demonstrated the potential of using LLMs to help solve multi-objective problems.
Several new methods were also proposed to enhance performance based on existing works. Wang et al.~\cite{wang_large_2024} proposed LLM-aided evolutionary search, where the LLM is responsible for generating only a 10\% of the offspring, while the remaining 90\% are produced through traditional evolutionary search operations. Their method has demonstrated promising outcomes in the realm of constrained multi-objective optimization~\cite{liang2023cmo}.
In another advancement, Brahmachary et al.~\cite{brahmachary_large_2024} proposed Language-model-based Evolutionary Optimizer (LEO), which is a specialized evolutionary optimizer that employs a dual-pool architecture, where one pool is dedicated to exploration and the other to exploitation, each with its own customized set of prompts tailored to the specific task.

Despite their differences among the aforementioned methods, they share a common feature: they operate within a traditional optimization framework (e.g. evolutionary algorithm), leveraging LLMs to supplant a crucial component within the iterative loop. This approach can be generally viewed as implementing a framework analogous to the evolutionary algorithm (EA) framework, where the LLM is employed as a key operator. For instance, in OPRO, the explicit maintenance of a list of best-found solutions is reminiscent of maintaining a population in EA with elite selection. Furthermore, the LLM's generation of new solutions based on this list can be seen as akin to mutation and crossover operations.

While numerous pioneering studies have explored the application of LLMs to numerical optimization, their study focuses on the methodology rather than evaluation, leaving the true effectiveness and characteristics of such methods largely unexplored.
Almost all existing works are built on the assumption that LLMs will revolutionize numerical optimization, without thoroughly evaluating its validity. 
This lack of critical evaluation has resulted in a significant knowledge gap. We aim to bridge this gap by providing a comprehensive understanding through rigorous empirical studies.

%% file: 3-experiment.tex
\section{Evaluation Settings\label{sec:settings}}

A summary of our evaluation process is illustrated in Fig.~\ref{fig:process}. In this section, we provide detailed evaluation settings, including model settings, prompt settings, problem settings, and procedure settings. 

\subsection{Model Settings}

To date, a considerable number of LLMs have been developed, 
each with distinct characteristics.
For a comprehensive evaluation of these models, it is crucial to cover a set of LLMs as diverse as possible. 
However, it is apparently unfeasible to cover them all due to the constraints in computational power, cost as well as accessibility. 
{Therefore, we initiate our evaluation with a group of baseline experiments and narrow down our focus to the top-performing models for further investigation.}

For baseline experiments, we dig into the capabilities of five frequently used LLMs, encompassing both closed and open-source variants, namely GPT-3.5, GPT-4, Gemini, LLaMA, and InternLM. 
These models are relatively up-to-date, popularly applied, and characterized by high parameter counts, as shown in Table~\ref{tab:llm_summary}.
The corresponding versions employed for each model are as follows: gpt-3.5-turbo-1106, gpt-4-0613, Gemini Pro, llama2-13b-chat, and interlm2-20b-chat. 
For the first three models, we use their online services to access the model, and for the last two models, we run them locally.

With a diverse set of tested LLMs, it is intractable to tune their hyper-parameters in our evaluation due to the huge computational cost. Therefore, we follow the commonly used setting.
Specifically, for close-sourced models (i.e. gpt-3.5-turbo-1106, gpt-4-0613, and Gemini Pro), we adhere to their default parameters, while for open-sourced models (i.e. llama2-13b-chat and interlm2-20b-chat), we use nucleus sampling~\cite{Holtzman2020The} with $p=0.95$ consistently.

\subsection{Prompt Settings}
In our evaluation, we always construct the prompts with three distinct parts: the initial part comprises a task description, followed by a list of the best solutions in history as the second part, and concluding with the task instruction in the third part. 
Considering the limited computational power, for each experiment setting, we repeat the experiments 5 times and report the average of the metric for that problem.
The selection of prompts significantly influences model behavior and performance, and pinpointing the best prompt for each model is barely feasible.
However, to ensure a fair comparison, it is necessary to tailor prompts properly to different LLMs.
Therefore, we curate a prompt pool, comprising prompts in which the task description and instruction parts are slightly different, and the part of best solutions in history is commonly shared among all models.
From this pool, we select the most effective prompt for each model. Additional details on these prompts are available in Appendix~\ref{sec:appendix} (a-g).

\subsection{Problem Settings}

\begin{figure*}
    \centering
    \begin{subfigure}{0.19\linewidth}
        \centering
        \includegraphics[width=\columnwidth]{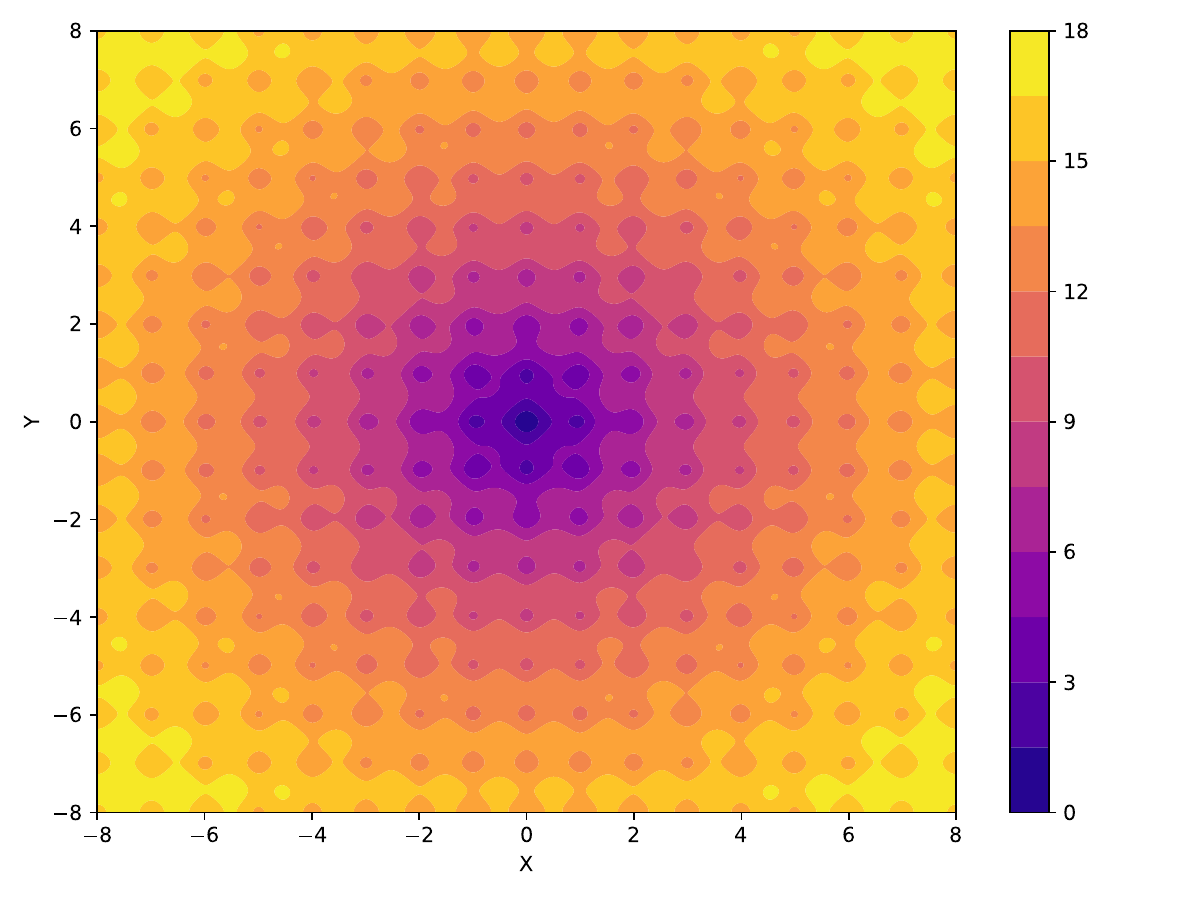}
        \caption{Ackley}
    \end{subfigure}
    \begin{subfigure}{0.19\linewidth}
        \centering
        \includegraphics[width=\columnwidth]{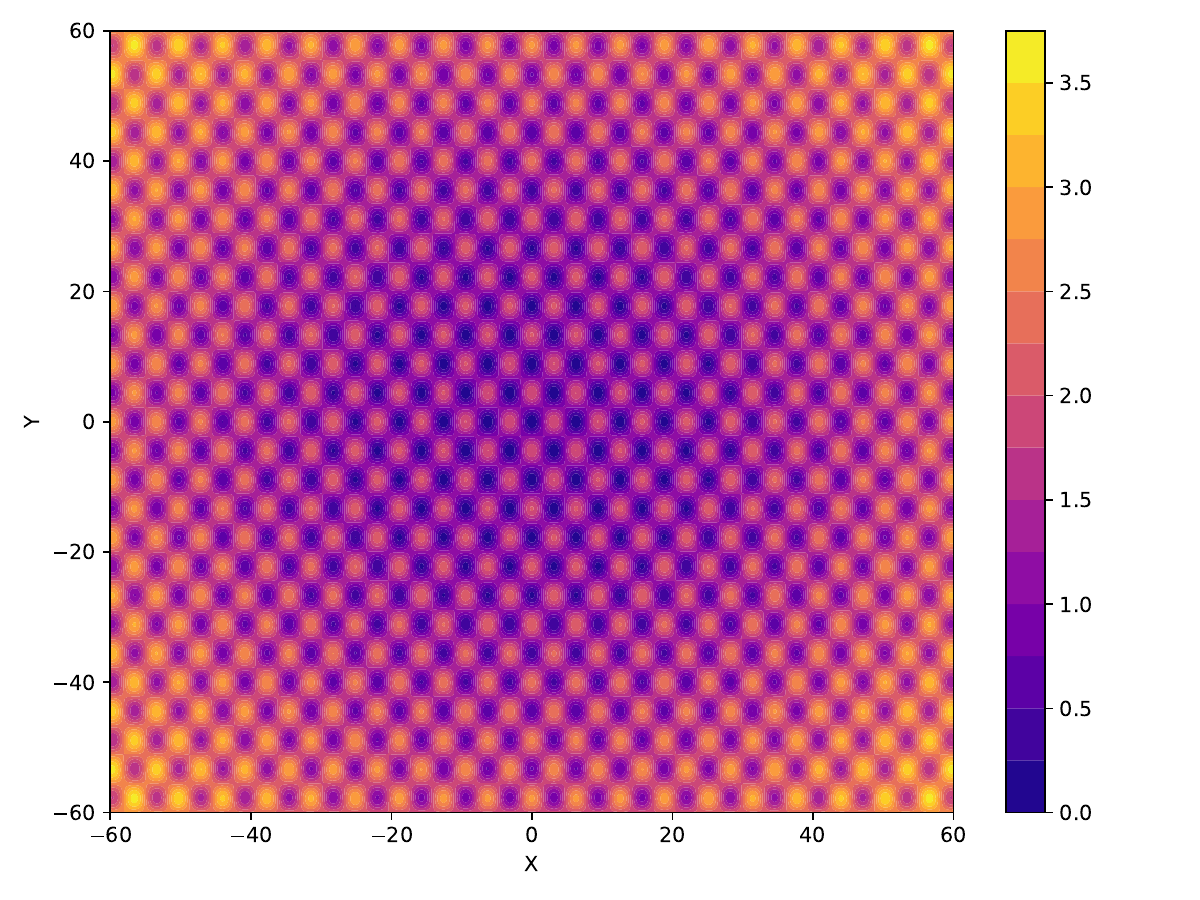}
        \caption{Griewank}
    \end{subfigure}
    \begin{subfigure}{0.19\linewidth}
        \centering
        \includegraphics[width=\columnwidth]{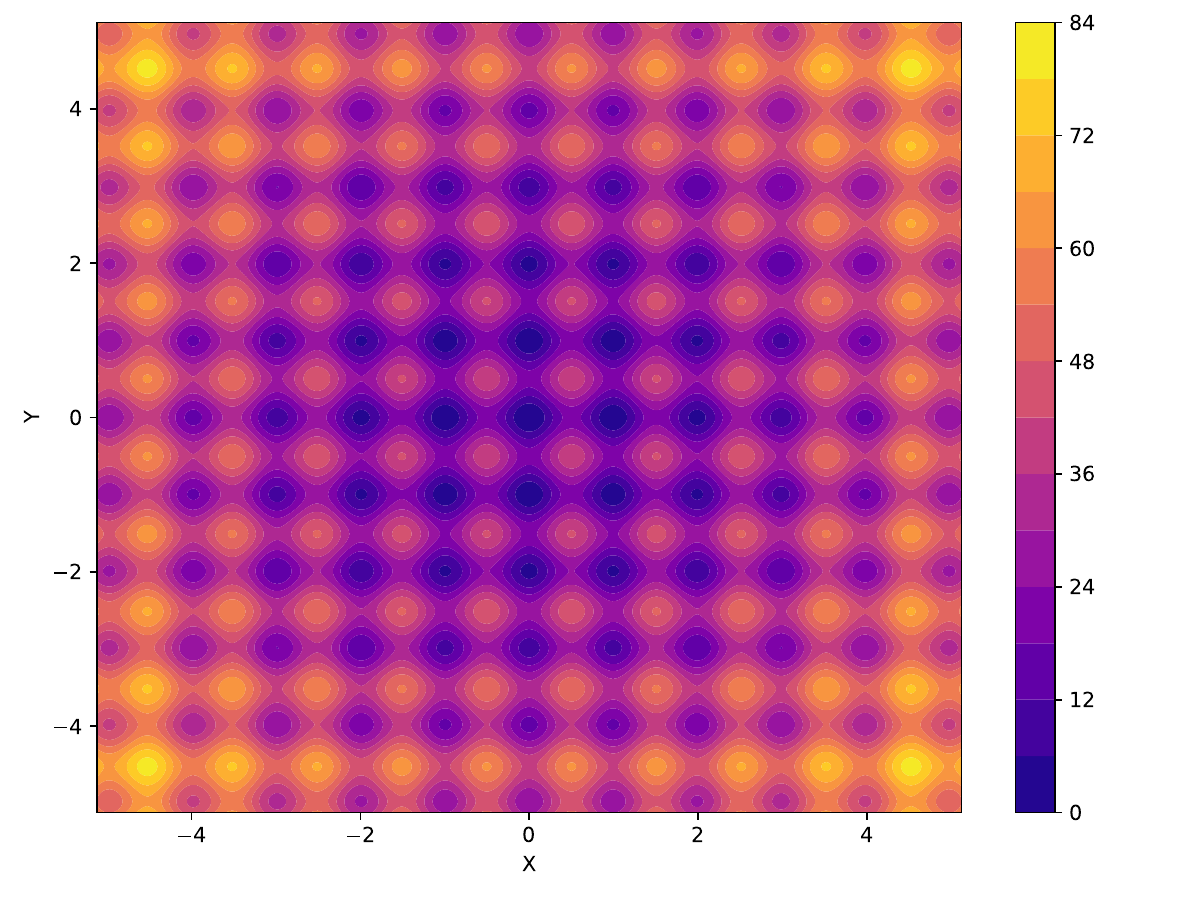}
        \caption{Rastrigin}
    \end{subfigure}
    \begin{subfigure}{0.19\linewidth}
        \centering
        \includegraphics[width=\columnwidth]{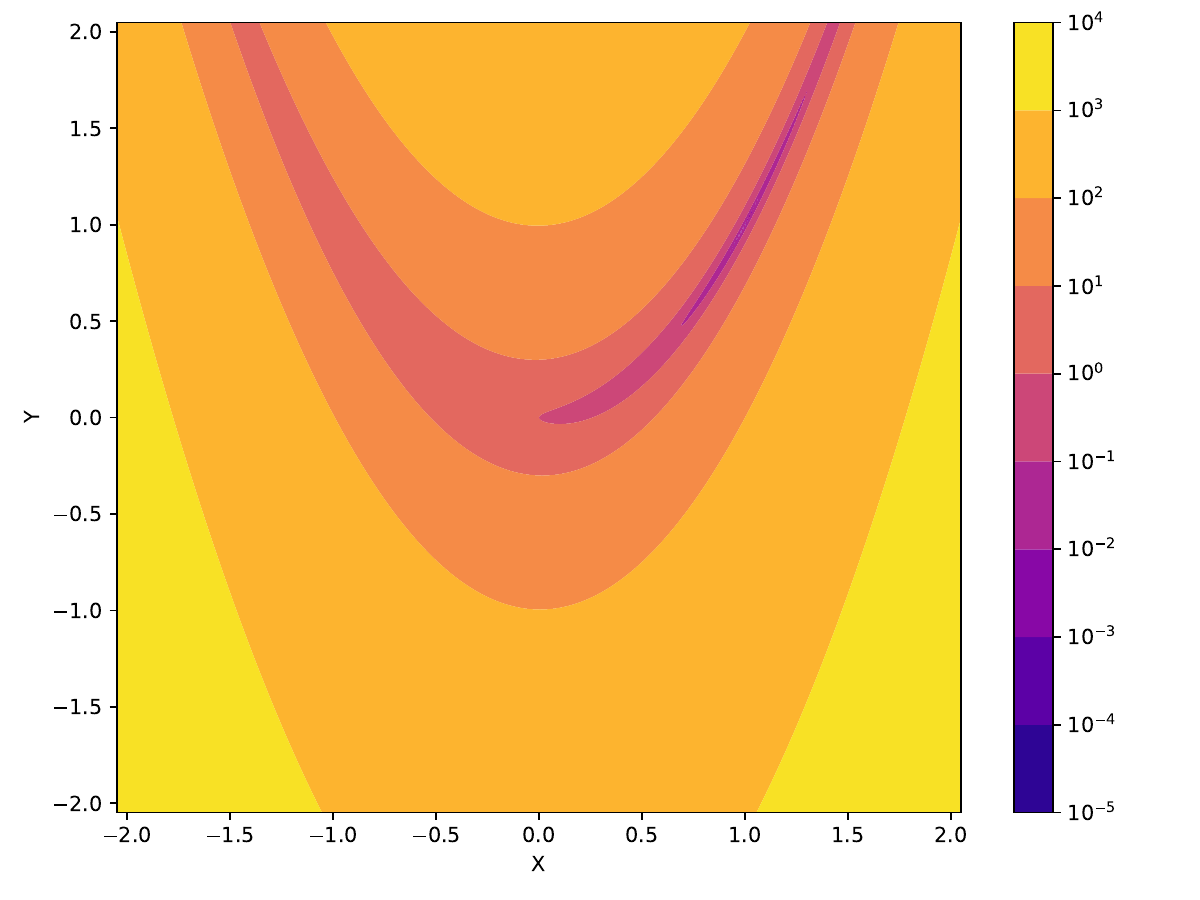}
        \caption{Rosenbrock}
    \end{subfigure}
    \begin{subfigure}{0.19\linewidth}
        \centering
        \includegraphics[width=\columnwidth]{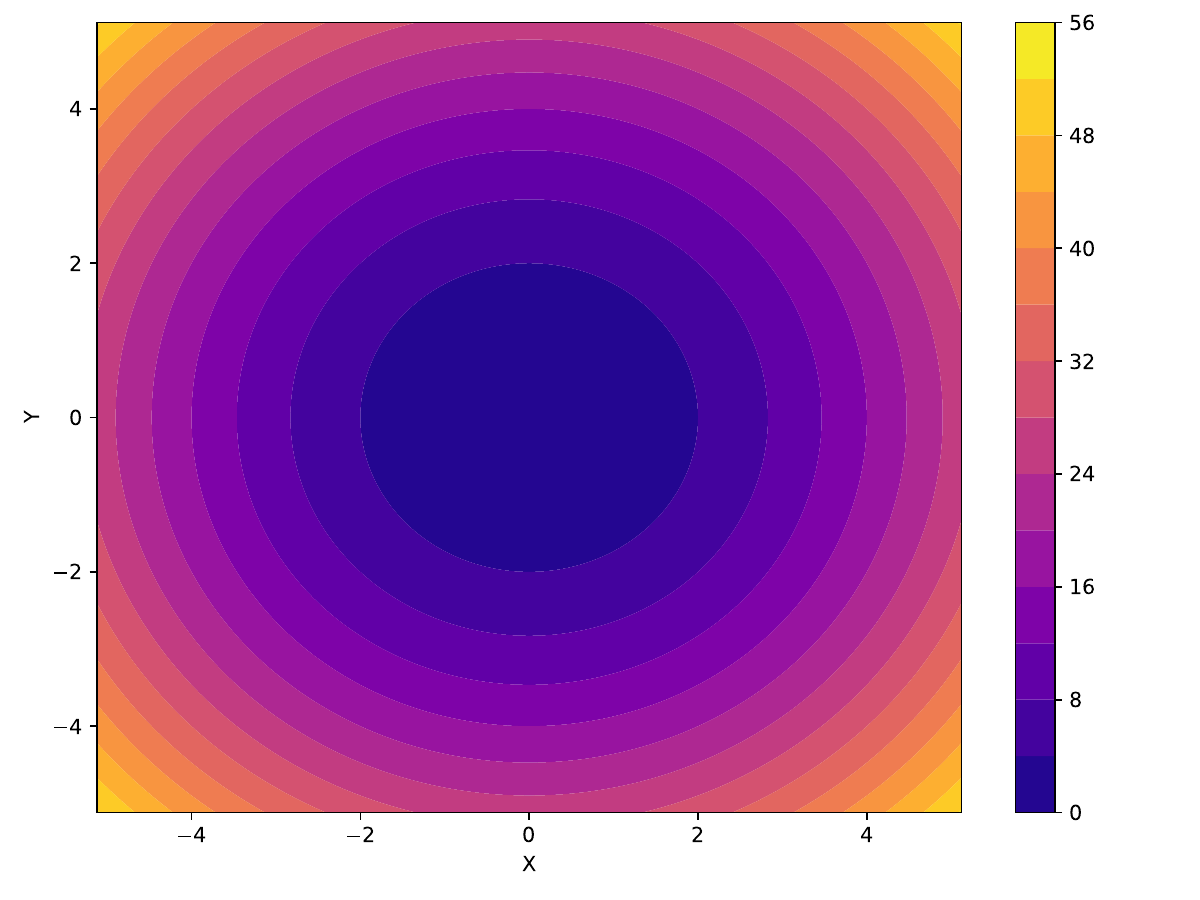}
        \caption{Sphere}
    \end{subfigure}
    \caption{The visualization of the landscape of numerical benchmark functions used in our investigation. To enhance clarity, the plotted region differs from the actual bounds utilized during the test. Additionally, the Rosenbrock function is depicted on a logarithmic scale to provide a more insightful representation.}
    \label{fig:landscape}
\end{figure*}

{We explore two types of benchmark problems in our study: discrete and continuous optimization problems. 
We choose the Travelling Salesman Problem (TSP)~\cite{gavish1978travelling,goyal2010survey} as a representative of the discrete problems. 
In TSP, given a list of cities and the distances between the cities, the task is to find the shortest possible route that visits each city exactly once before returning to the origin city. 
The continuous problems we use for investigation include Ackley, Griewank, Rastrigin, Rosenbrock, and Sphere~\cite{jamil2013literature}.
These problems involve functions that take multiple real-valued inputs and produce a single real-valued output, which represents the cost we aim to minimize. 
Typically, such functions have a single global minimum point and multiple local minima. The goal is to find the optimal input values that minimize the function's output. Specifically, a visualization of the landscape of the functions we have chosen is given in Fig.~\ref{fig:landscape}. }
We select these problems as benchmarks due to their longstanding status in the optimization field as classic challenges.

\subsection{Procedure Settings}

With the aforementioned settings in terms of the model, prompt, and problem, we further construct an optimization procedure for our investigation.
It is built to cover the essential elements identified in prior studies~\cite{yang_large_2023,guo_towards_2023,meyerson_language_2023,liu_large_single_2023,lange_large_2024,huang_how_2024,liu_large_multi_2023,nasir_llmatic_2023,wang_large_2024,brahmachary_large_2024}, such as the manual maintenance of a solution set and the generation of new solutions from LLMs. Specifically, it goes as follows: 
\begin{enumerate}
    \item Generate $n$ random solutions as the initial solution set;
    \item Prompt the LLM with the problem-specific prompt alongside the top $n$ solutions to generate $m$ new solutions;
    \item Update the top $n$ solutions;
    \item Repeat Steps 2) and 3) until the stop condition is met.
\end{enumerate}
It is noteworthy that in Step 2) in order to ensure more than one solution to be generated,
we query the model with the same prompt multiple times in a single iteration.
For each response, we validate the solution format to ensure its correctness. If the LLM generates an invalid solution, we will resend the request until a valid solution is obtained or the model exceeds a predetermined number  %\bh{we set to 100}
of failed attempts {(we set to 100 in our implementation)}.

\section{Investigation and Analysis \label{sec:Investigation}}

{In this section, we begin with a baseline performance evaluation of all the tested models over simple tasks that cover both discrete and continuous problems. 
Based on the baseline performance, we select the top performers for subsequent analysis. During this stage, we first focus on analyzing the very basic properties that are inherent in existing black-box optimization algorithms, and then go deeper to investigate the extent to which LLMs can leverage their extensive knowledge base to formulate effective heuristics during the optimization process.

\subsection{Baseline Performance \label{sec:baseline}}

We first examine the performance of various LLMs in the realm of black-box optimization. Specifically, we evaluate their performance from two aspects: the solution quality and the proficiency in maintaining a valid output format. The latter is represented as the incidence of invalid results.
Unlike traditional optimization algorithms, LLMs may occasionally generate outputs that do not conform to the required format. While this can be mitigated by retrying and allowing the LLM to generate the result again, a high frequency of generating invalid outputs can result in elevated costs. Moreover, it signifies a potential deficiency in the LLM's capacity in understanding the problem and the provided instructions. 
In our experiments, considering performance variations among models as well as resource constraints, we place restrictions on the number of allowable retrials in the case of a model producing invalid results; once the allocated retries are fully utilized, the optimization process is deemed unsuccessful and terminated.

For the TSP, we use five different settings with varying numbers of cities: TSP-10, TSP-15, TSP-20, TSP-25, and TSP-30, including 10, 15, 20, 25, and 30 cities, respectively. 
In each setting, we randomly generate the coordinates of the cities on a 2D plane, with both the x and y coordinates being integers within the range of 0 to 100. 
The variation in the number of cities not only challenges the models' optimization capabilities but also tests their proficiency in generating valid permutations. 
To measure how well the final solution is, we adopt the Concorde TSP solver~\cite{applegate2006concorde} to compute the exact solution for each setting and further compute the performance gap with respect to the tested LLMs.

For all continuous numerical benchmark functions, i.e. Ackley, Griewank, Rastrigin, Rosenbrock, and Sphere,
we use a fixed problem dimension of 2 and employ default boundaries for the 2D problem settings.
On one hand, this choice is for simplicity, as we try to avoid overly challenging conditions and maintain a balanced difficulty level for this baseline test. 
On the other hand, the utilization of 2D problem settings enables us to visually comprehend the landscape of these numerical benchmark functions. 
Finally, the minimum function output is reported to assess the optimization quality.

During the evaluation, we employ the same random number seed across all problem settings and initial solution generation, ensuring that all models commence with identical initial solutions. 
It is worth noting that, for TSP, this implies that all models are confronted with the same city arrangement, establishing a consistent and fair basis for comparison.
All the results for this evaluation are summarized in Table~\ref{tab:baseline_performance}.

\begin{table*}[]
    \centering
    \caption{Comparison of the performance of baseline models. For discrete benchmarks, we report the average optimization gap and failure rate (statistics before and after the slash, respectively). For continuous benchmarks, we report the average of the best fitness and the average number of retrials (statistics before and after the slash, respectively). {The `-' denotes the scenario in which the model exceeds the maximum attempts.}}
    \begin{tabularx}{\linewidth}{X X X X X X X}
        \toprule
        \multicolumn{2}{c}{Problems} & \multicolumn{5}{c}{Large Language Models} \\
        Problem Type & Problem Settings & GPT-3.5 & GPT-4 & Gemini & LLaMA & InternLM \\
        \toprule
        \multirow{5}{*}{Discrete} & TSP-10 & 0\% / 0.2      & \textbf{0.00\%} / \textbf{0} & 0\% / 0.8   & 35.21\% / 5.25  & - / - \\
                                  & TSP-15 & 6.01\% / 3.2   & \textbf{0.28\% / 1.2}     & 4.69\% / 16.8  & 87.07\%   / 15        & - / - \\
                                  & TSP-20 & 30.69\% / 10.8 & \textbf{0.88\% / 2.6}     & 4.21\% / 32.75 & 141.64\%  / 19.33    & - / - \\
                                  & TSP-25 & 31.20\% / 24.2 & \textbf{3.38\% / 10.8}    & - / -          & - / - & - / - \\
                                  & TSP-30 & - / -          & \textbf{11.01\% / 5.6}    & - / -          & - / - & - / - \\
        \midrule
        \multirow{5}{*}{Continuous} & Ackley & 9.08 / 0 & \textbf{7.40 / 0} & 11.34 / 0 & 16.91 / 0 & - / - \\
                                  & Griewank & 2.20 / 0 & \textbf{0.33 / 0} & 5.71 / 0 & 11.91 / 0 & - / - \\
                                 & Rastrigin & 2.43 / 0 & \textbf{1.39 / 0} & 2.57 / 0 & 9.36 / 0 & - / - \\
                                & Rosenbrock & 2.77 / 0 & \textbf{1.74 / 0} & 1.96 / 0 & 6.73 / 0 & - / - \\
                                    & Sphere & 1.14 / 0 & \textbf{0.0 / 0}  & 0.00 / 0 & 3.23 / 0 & - / - \\
        \bottomrule
    \end{tabularx}
    \label{tab:baseline_performance}
\end{table*}

According to the results shown in Table~\ref{tab:baseline_performance}, in the TSP task, none of the tested LLMs consistently generates the correct format throughout the test. 
Moreover, the likelihood of models generating invalid results increases as the number of cities grows.
This behavior is expected.
Unlike traditional algorithms that have specific operators to deal with permutation, LLMs directly work with permutation in its string representation, which means they suffer a greater challenge to keeping a valid permutation as the number of cities grows.
Comparatively, continuous benchmark problems only require the output to be within a certain range with an upper bound and a lower bound.
This means it is easier for the models to consistently produce valid output on these problems.

Overall, different LLMs show varying effectiveness on addressing optimization problems.
Notably, GPT-4 demonstrates unparalleled performance across all tasks, surpassing both GPT-3.5 and Gemini; LLaMA2 falls behind in terms of both optimization outcomes and capacity to generate valid results; InternLM ranks the last, primarily due to the verbosity of its output, consistently producing excessive output that frequently disrupts the specified output format requirements. Appendix~\ref{sec:appendix} (h) (i) provides two examples of invalid output.

Due to limitations in computational power and significant performance differences among various LLMs, we only study the optimization properties of the top-performing LLMs in the subsequent research. 
Specifically, GPT-3.5, GPT-4, and Gemini are selected as the models to be further investigated due to their ability to consistently produce valid output and achieve relatively good optimization results.

\begin{figure}
    \centering
    \includegraphics[width=\columnwidth]{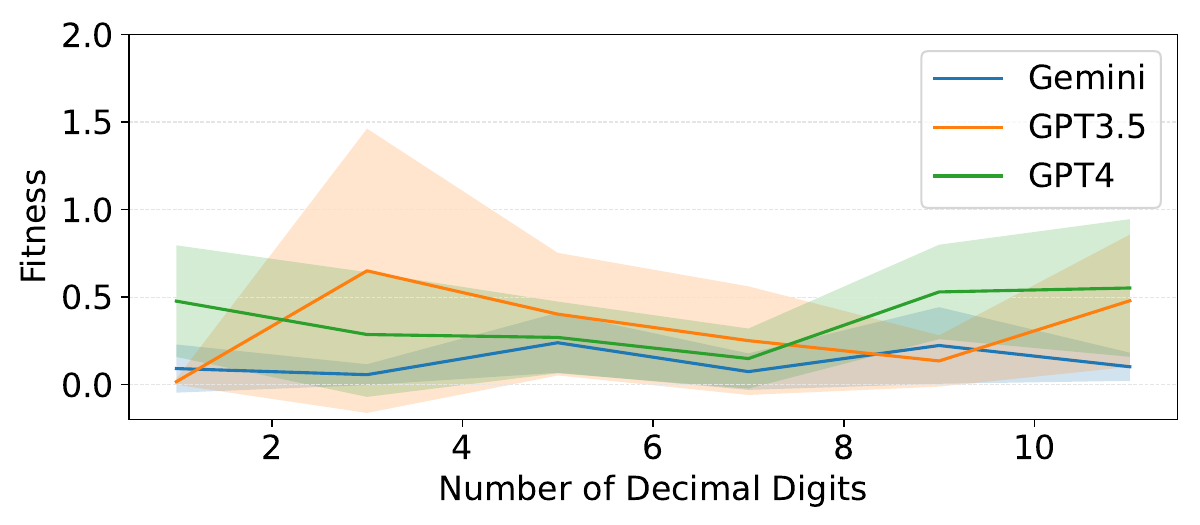}
    \caption{Evaluation results of LLMs' capacity in comprehending string-represented numbers. We manipulate the number of decimal digits in the input to control its numerical precision.}
    \label{fig:llm_number_understanding}
\end{figure}

\begin{figure}
    \centering
    \includegraphics[width=\columnwidth]{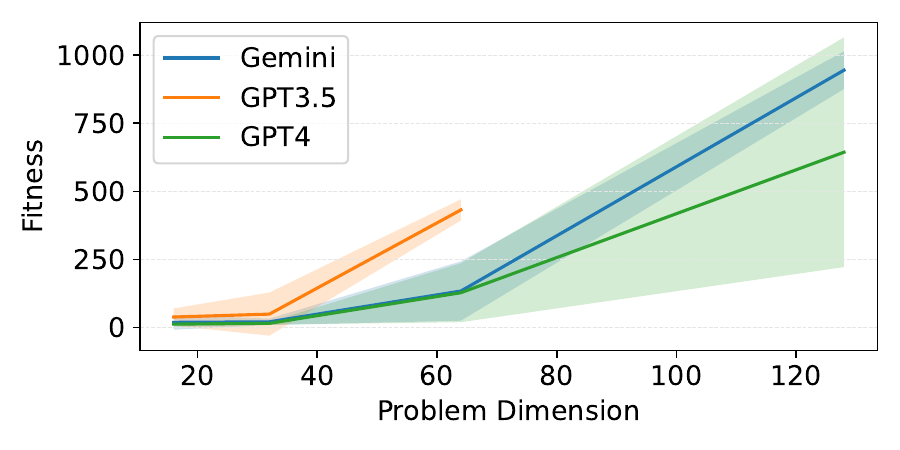}
    \caption{Evaluation results of LLMs' scalability on problem dimensions. We scale the number of dimensions from 16 to 256 using a sphere function.
    }
    \label{fig:llm_scalability}
\end{figure}

\subsection{Basic Properties \label{sec:basic}}

In this subsection, we investigate whether LLMs manifest the properties that typical optimizers would have.

\noindent\textbf{Understanding of Numerical Values.}
\label{subsec:number}
Traditional optimization algorithms directly manipulate numerical values through fundamental operations, e.g., addition and multiplication, on binary representations of numbers. Such an inherent capability enables them to handle higher numerical precisions with ease. This is, however, not guaranteed with LLMs, which deal with string data and always encode numbers as tokens.
In this part, we evaluate the competency of LLMs in handling numerical values.

Our evaluation is conducted using a sphere function, chosen for its inherent simplicity. We believe that this function will not overly challenge the optimization ability of the model, allowing us to focus primarily on assessing its numerical understanding capability.
In order to reduce the possibility of the model guessing the optimum solution (i.e., the point $(0,0)$), we apply a small random shift of a real value ranging from -0.1 to 0.1 in both $x_1$ and $x_2$ directions. The shift is kept small enough to avoid transforming the problem into a more complex one.
With the optimum point not set to be at an integer location, to get better fitness, the model needs to output the solution that is as close to the random shift as possible.
We then control the precision of the input value given to the model by controlling the number of decimal digits in the input format.
Details about how the number is formatted can be found in Appendix~\ref{sec:appendix} (d). In ideal cases, since the random shift is a high-precision real value, and we put a hard limit on the number of decimal digits of the input numerical value in the prompt, a lower precision will prevent the model from approaching the random shift, and vice versa.
We use the same random number seed for all the precision settings in our experiments, i.e., using the same random shifts across different precision settings, for a fair comparison.

We plot all the experimental results in Fig.~\ref{fig:llm_number_understanding}.
It can be seen that LLMs do not consistently benefit from increased precision of the input numbers, and may even exhibit declined performance with additional decimal digits.

In particular, Gemini is the most stable model, followed by GPT-4. However, even the best-performing model Gemini exhibits a certain degree of fluctuation, with no guarantee of non-decreasing performance.
This behavior is unusual, as empirically, increasing input precision tends to either enhance performance or have little impact on the program. We attribute these results to LLMs not fully understanding the input numerical values and being sensitive to the given prompt, which also changes with increased precision.
Consequently, this makes LLMs less suitable for tasks requiring high precision.
Additionally, more decimal digits in the input format increase the number of tokens fed to the LLM, leading to higher computational costs. Therefore, we suggest that setting a smaller number of decimal digits is more suitable for addressing optimization problems with LLMs.

\noindent \textbf{Scalability on Problem Dimensions.}
\label{subsec:scale}
We extend the benchmark sphere function to $N$ dimensional settings, so as to evaluate the scalability of LLMs on problem dimensions. 
In this study, we continue to employ a small random shift to the optimal solution. This is because LLMs have a relatively high tendency to output a zero vector, which is the optimum of this problem in all $N$ dimensional settings, thus contradicting the intended purpose of this test. In this problem, each dimension operates independently and, therefore, can be optimized independently. Thus,  scaling the problem dimension will not largely harm the optimizer's performance. Specifically, we systematically escalate the problem dimension of the sphere function exponentially, progressing from 16 to 256 dimensions, while the optimizer runs for a fixed 100 iterations.

We present the experimental results in Fig.~\ref{fig:llm_scalability}.
From this figure, it is evident that as the number of dimensions exponentially increases, there is a consistent decline in performance.
Even with the initial setting of 16 dimensions, the three tested models exhibit a failure to find the optimum solution, in contrast to their success in doing so during the baseline experiments conducted with a dimensionality of 2 (Section~\ref{sec:baseline}).
Among the three models, GPT-3.5 suffers the most severe drop in performance from the scaling of the problem dimension.
When the problem dimension is scaled 4 times, i.e., from 16 dimensions to 64 dimensions, its performance degrades more than 10 times. 
Also, it is worth noting that the scalability of LLMs is constrained by the maximum context length. 
Specifically, GPT-3.5 experiences a complete failure at 128 dimensions, while GPT-4 will not surpass this threshold, i.e., 128 dimensions, due to the context length limit. Specifically, GPT-3.5 experiences a complete failure at 128 dimensions, while GPT-4, with its greater context length, reaches 128 dimensions but fails to scale further due to the context length limit. Such a limit of context length poses a significant constraint on the practical use of LLMs in optimization, directly impacting their scalability.

\noindent \textbf{Adaptability to Diverse Problem Scenarios.}
\label{subsec:shift}
\begin{figure*}
    \centering
    \begin{subfigure}{0.24\linewidth}
        \centering
        \includegraphics[width=\columnwidth]{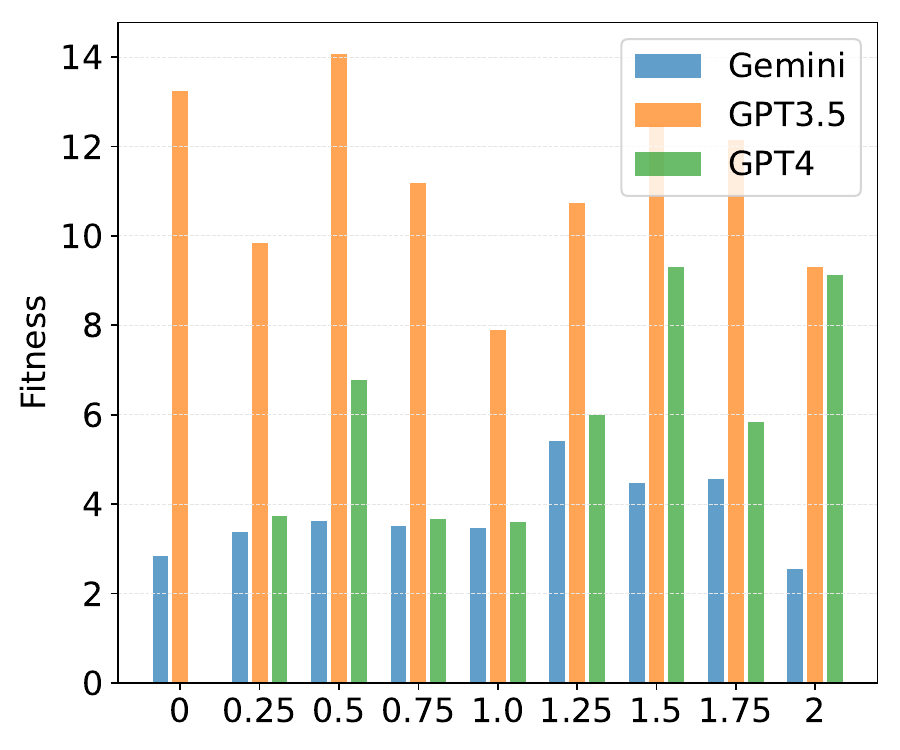}
        \caption{Ackley}
    \end{subfigure}
    \begin{subfigure}{0.24\linewidth}
        \centering
        \includegraphics[width=\columnwidth]{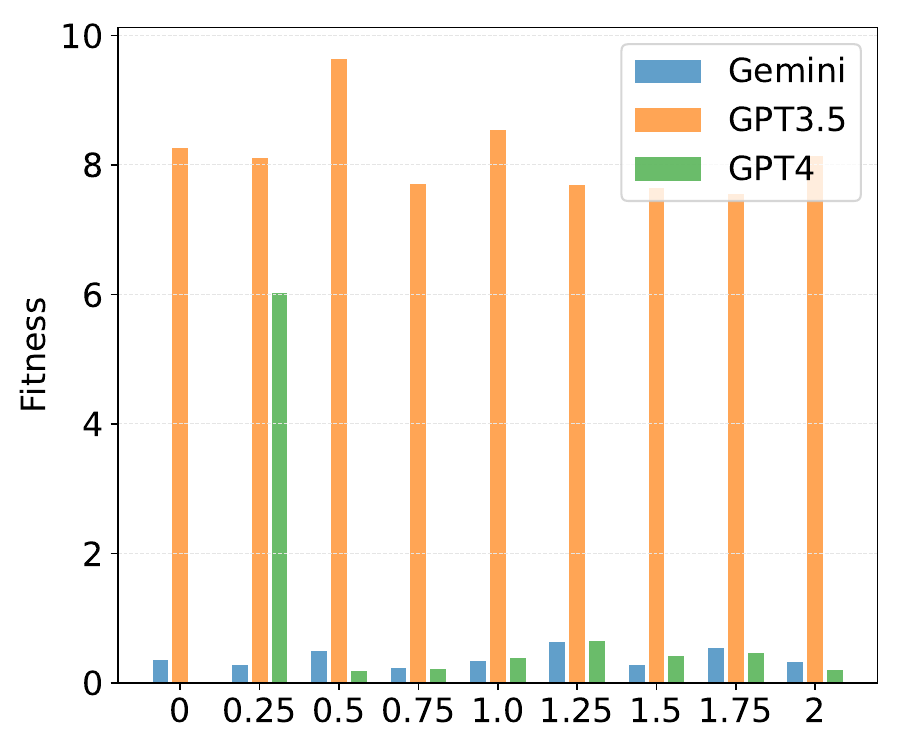}
        \caption{Griewank}
    \end{subfigure}
    \begin{subfigure}{0.24\linewidth}
        \centering
        \includegraphics[width=\columnwidth]{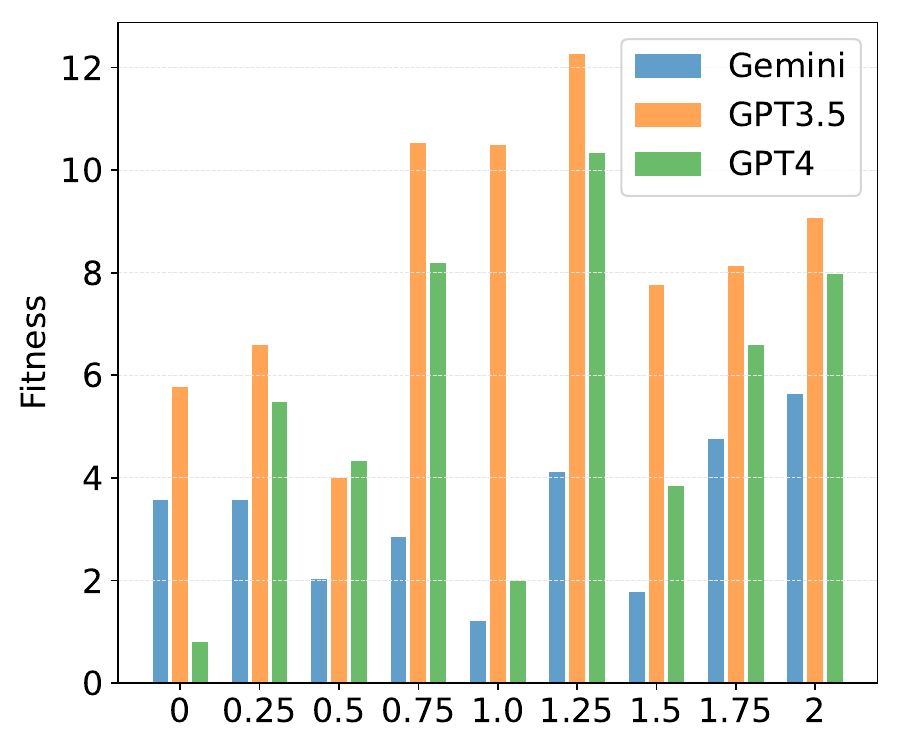}
        \caption{Rastrigin}
    \end{subfigure}
    \begin{subfigure}{0.24\linewidth}
        \centering
        \includegraphics[width=\columnwidth]{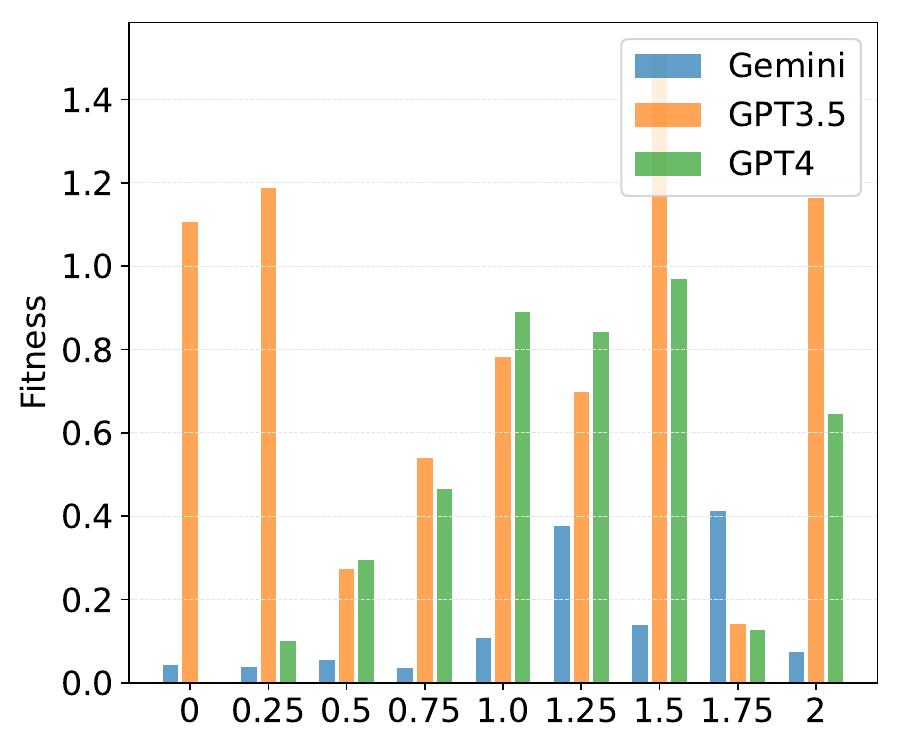}
        \caption{Sphere}
    \end{subfigure}
    \caption{Evaluation results of LLMs' adaptability to different continuous optimization problems. Performance fluctuates drastically when the given input undergoes a shift. Notably, the tested models exhibit distinct responses to these shifts. The average performance over 5 runs is reported, as described in Section III.B.}
    \label{fig:llm_shift}
\end{figure*}
In this subsection, we investigate the performance of LLMs in terms of their adaptability to diverse problems. To this end, we create a set of problem variants by modifying the original problem and evaluating the performance of LLMs on this set of problems.
Specifically, we add a constant vector, that is randomly generated with the magnitude varying from 0 to 2 in all dimensions, to the inputs of our benchmark function. Our test includes four distinct benchmark functions, i.e., Ackley, Griewank, Rastrigin, and Sphere in their 2-dimensional forms.
We omit the Rosenbrock function for its asymmetric nature, as in this test, the direction of the shift is a significant factor besides the magnitude, making Rosenbrock less ideal for this specific testing condition.

We provide the results in Fig.~\ref{fig:llm_shift}.
It is apparent that all LLMs are influenced to some extent by the shift, indicating a lack of shift-invariance, a characteristic commonly possessed by traditional optimizers. 
Among them, GPT-4, the best performing model in the baseline test, continues to lead when no shift is applied, achieving close to optimal results across all four benchmark functions. 
However, even a little shift could make GPT-4 miss the optimum solution. 
Among the three assessed models, Gemini exhibits relatively less susceptibility to the shift and displays minimal fluctuations in performance, although it is not the best at solving the original optimization problems.

\noindent \textbf{Balancing Exploration and Exploitation.}
\label{subsec:balance}
Balancing exploration and exploitation is a crucial aspect of optimization strategies, representing the delicate equilibrium between two fundamental objectives. Exploration entails searching the solution space to discover new and potentially superior solutions, while exploitation involves refining established solutions to maximize immediate gains.
Striking the right balance between them is essential in optimizing processes.
An excessive focus on exploration may overlook promising solutions, while an overemphasis on exploitation risks premature convergence on suboptimal outcomes.

In this test, we investigate  LLMs' proficiency in achieving this delicate balance. 
We visualize the sampling behavior of the three LLMs, i.e. Gemini, GPT-3.5, and GPT-4, on a 2-dimensional sphere function. 
We inform the tested LLMs of the top 16 solutions in the prompts and keep the remaining areas unexplored.
We use two different sets of top 16 solutions. 
As Gemini, GPT-3.5, and GPT-4 are all closed-source LLMs with undisclosed output distributions, we employ the Monte Carlo method to estimate their output distributions, which iteratively prompts LLMs to generate new solutions using the exact same prompt.
For each set, we apply a sample size of 1000, which is empirically sufficient for estimating the output distribution for the three tested models.

\begin{figure*}
    \centering
    \begin{subfigure}{0.24\linewidth}
        \centering
        \includegraphics[width=\columnwidth]{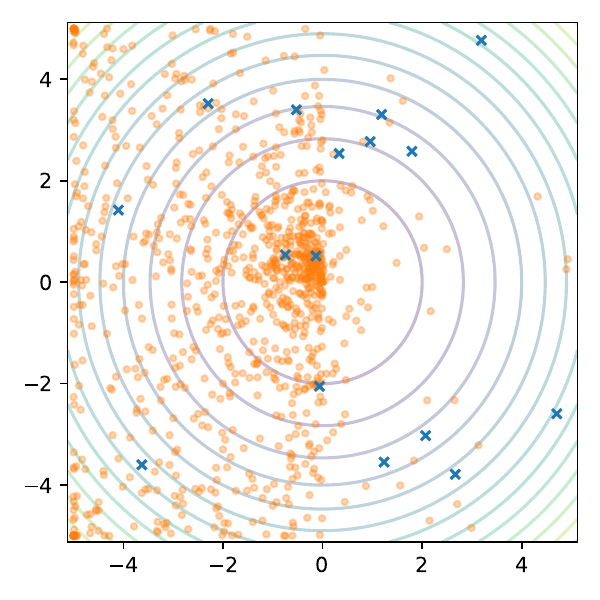}
        \caption{Gemini - No shift - Ascend}
    \end{subfigure}
    \begin{subfigure}{0.24\linewidth}
        \centering
        \includegraphics[width=\columnwidth]{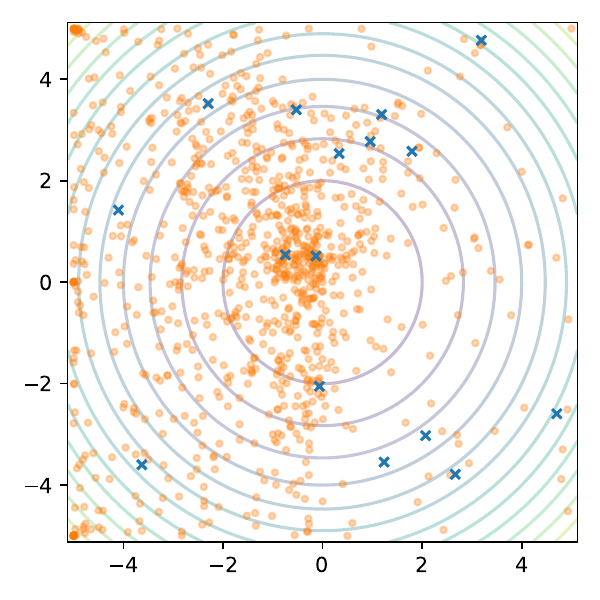}
        \caption{Gemini - No shift - Descend}
    \end{subfigure}
    \begin{subfigure}{0.24\linewidth}
        \centering
        \includegraphics[width=\columnwidth]{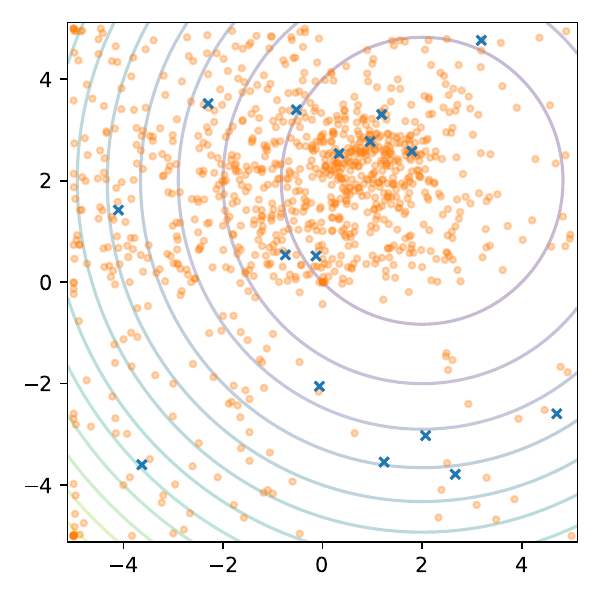}
        \caption{Gemini - Shifted - Ascend}
    \end{subfigure}
    \begin{subfigure}{0.24\linewidth}
        \centering
        \includegraphics[width=\columnwidth]{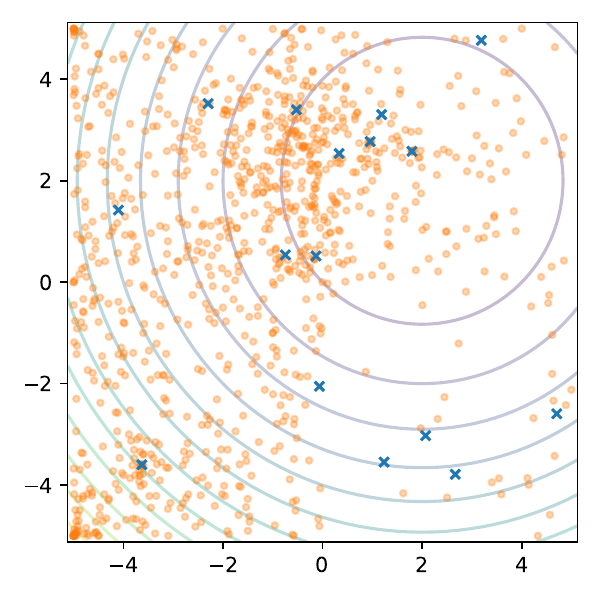}
        \caption{Gemini - Shifted - Descend}
    \end{subfigure}

    \begin{subfigure}{0.24\linewidth}
        \centering
        \includegraphics[width=\columnwidth]{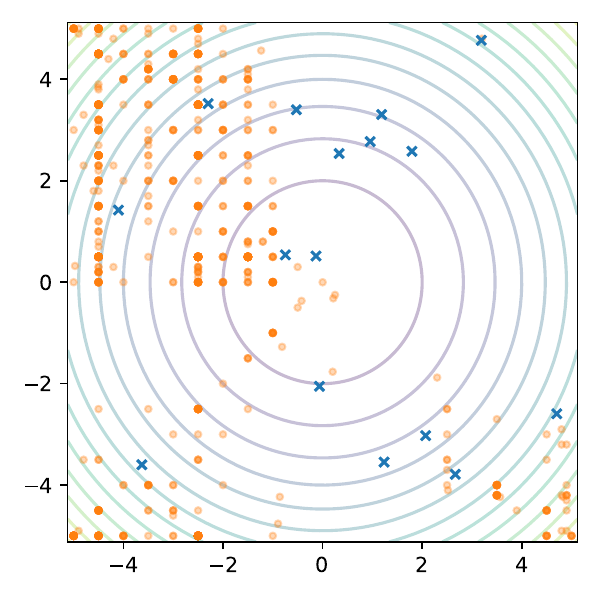}
        \caption{GPT-3.5 - No shift - Ascend}
    \end{subfigure}
    \begin{subfigure}{0.24\linewidth}
        \centering
        \includegraphics[width=\columnwidth]{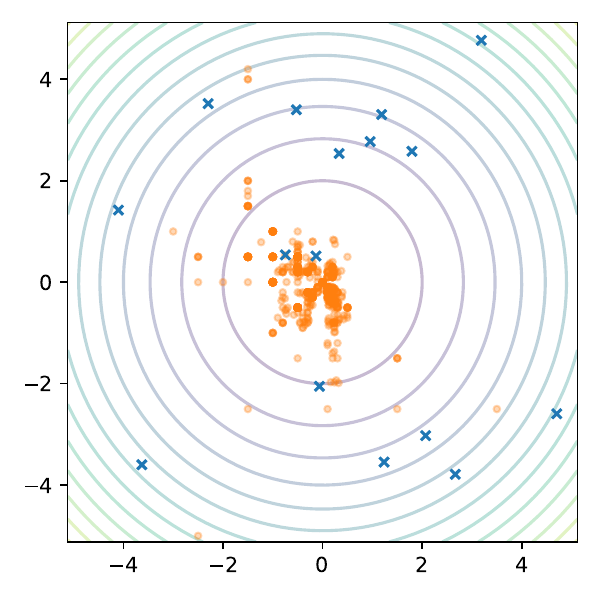}
        \caption{GPT-3.5 - No shift - Descend}
    \end{subfigure}
    \begin{subfigure}{0.24\linewidth}
        \centering
        \includegraphics[width=\columnwidth]{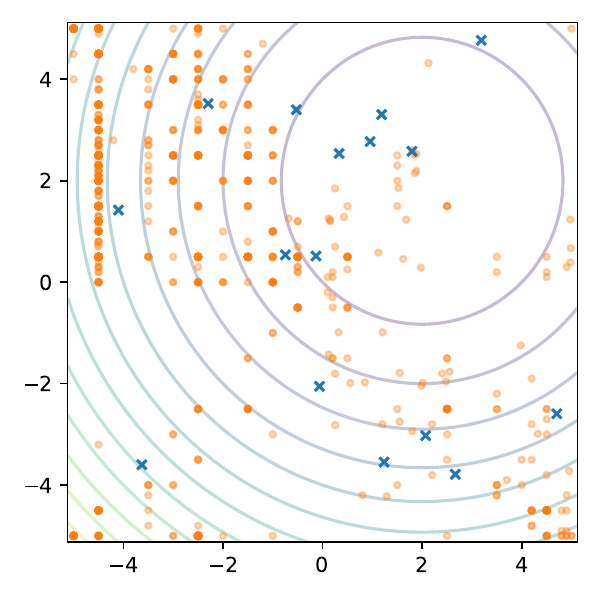}
        \caption{GPT-3.5 - Shifted - Ascend}
    \end{subfigure}
    \begin{subfigure}{0.24\linewidth}
        \centering
        \includegraphics[width=\columnwidth]{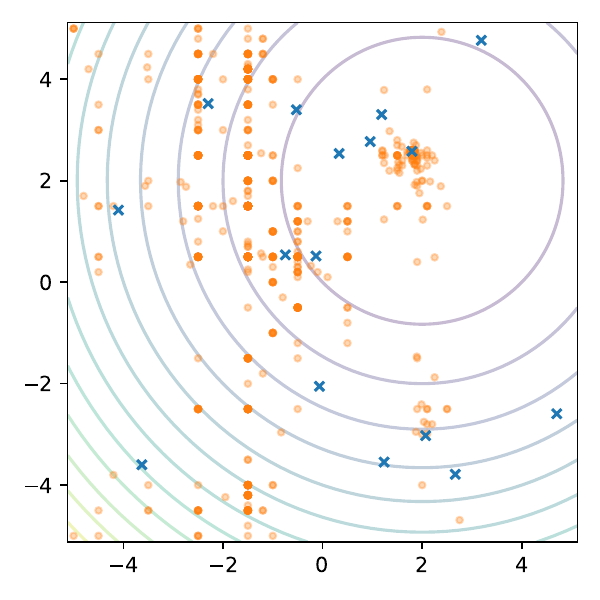}
        \caption{GPT-3.5 - Shifted - Descend}
    \end{subfigure}

    \begin{subfigure}{0.24\linewidth}
        \centering
        \includegraphics[width=\columnwidth]{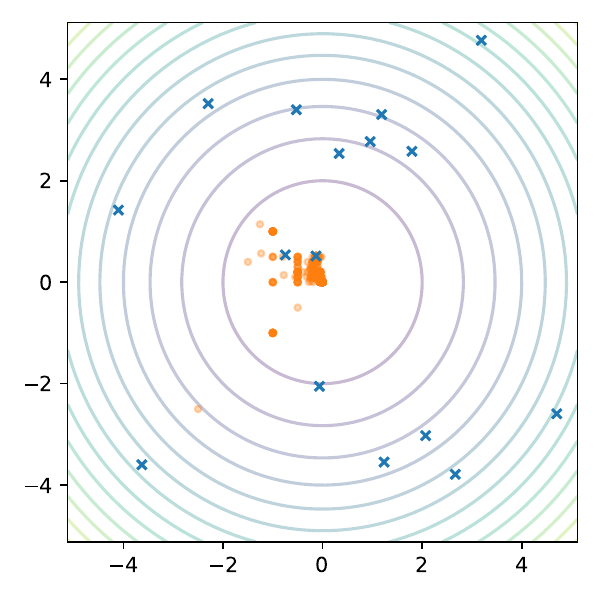}
        \caption{GPT-4 - No shift - Ascend}
    \end{subfigure}
    \begin{subfigure}{0.24\linewidth}
        \centering
        \includegraphics[width=\columnwidth]{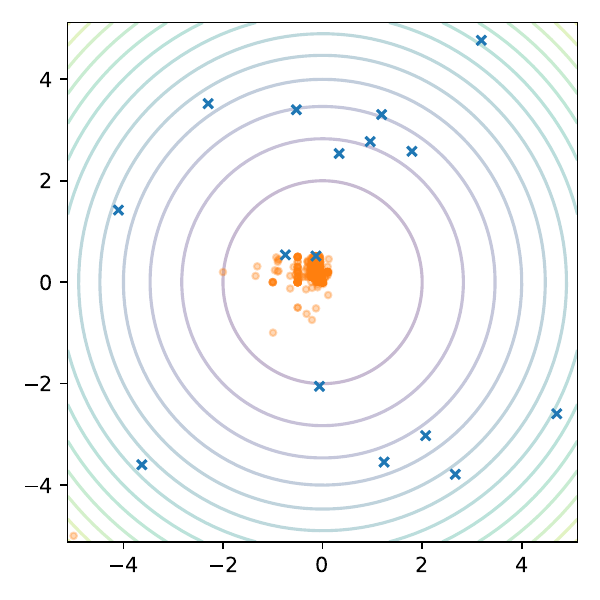}
        \caption{GPT-4 - No shift - Descend}
    \end{subfigure}
    \begin{subfigure}{0.24\linewidth}
        \centering
        \includegraphics[width=\columnwidth]{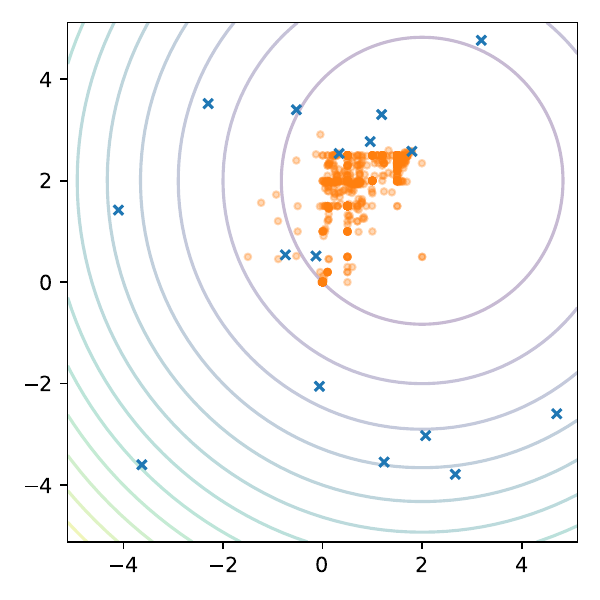}
        \caption{GPT-4 - Shifted - Ascend}
    \end{subfigure}
    \begin{subfigure}{0.24\linewidth}
        \centering
        \includegraphics[width=\columnwidth]{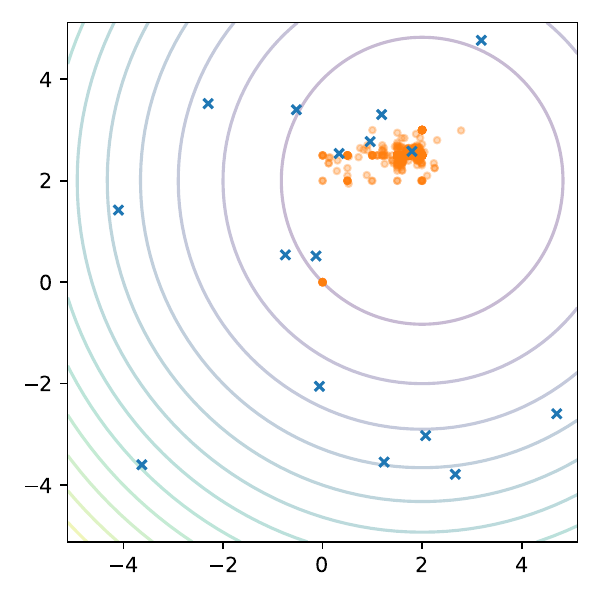}
        \caption{GPT-4 - Shifted - Descend}
    \end{subfigure}
    \caption{An illustration of LLMs' generation behavior with the 1st set of top-16 solutions. Background contours stand for the landscape of the sphere function; blue crosses are explored points (i.e., the top-16 solutions) given to the LLM as a context; semi-transparent orange dots denote the points the LLM chooses to explore next.}
    \label{fig:llm_behavior_1}
\end{figure*}

\begin{figure*}
    \centering
    \begin{subfigure}{0.24\linewidth}
        \centering
        \includegraphics[width=\columnwidth]{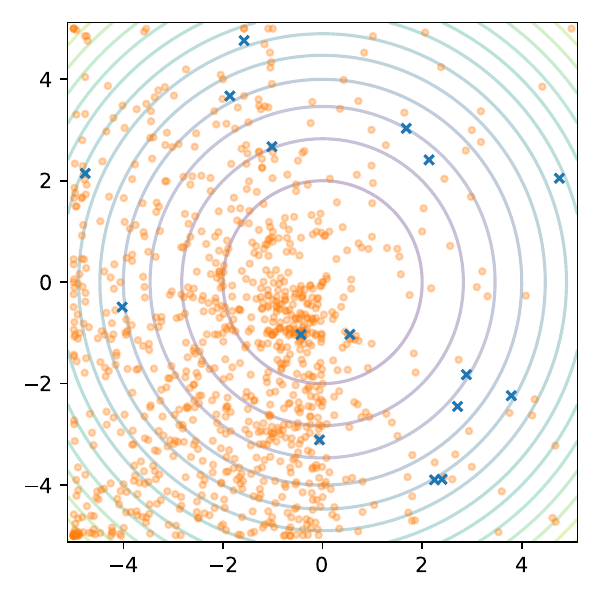}
        \caption{Gemini - No shift - Ascend}
    \end{subfigure}
    \begin{subfigure}{0.24\linewidth}
        \centering
        \includegraphics[width=\columnwidth]{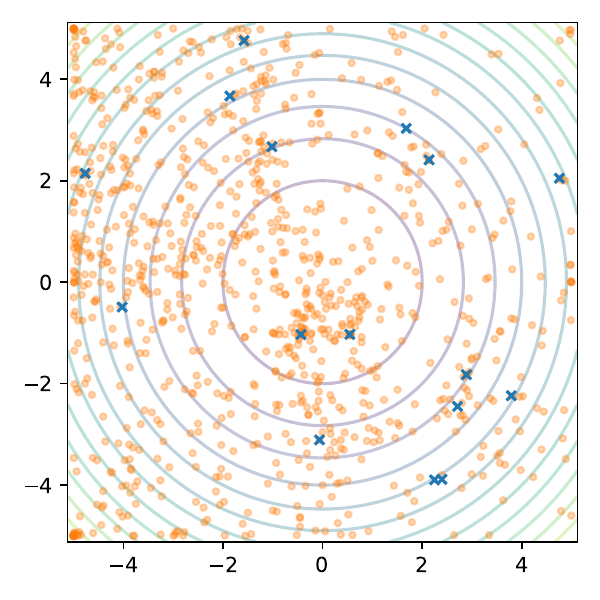}
        \caption{Gemini - No shift - Descend}
    \end{subfigure}
    \begin{subfigure}{0.24\linewidth}
        \centering
        \includegraphics[width=\columnwidth]{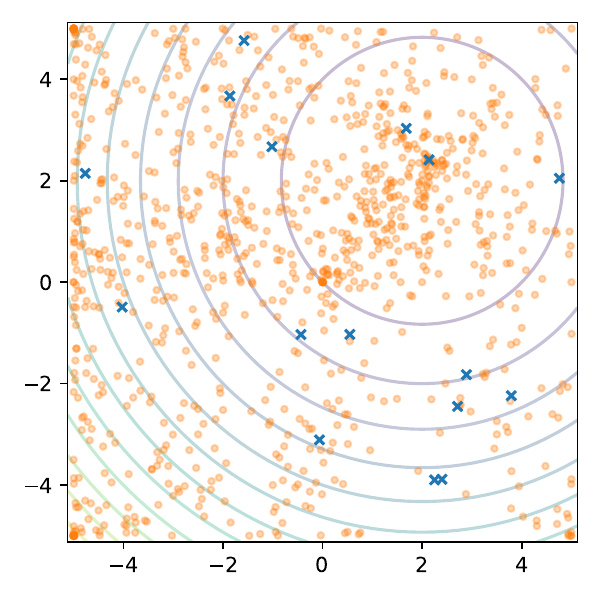}
        \caption{Gemini - Shifted - Ascend}
    \end{subfigure}
    \begin{subfigure}{0.24\linewidth}
        \centering
        \includegraphics[width=\columnwidth]{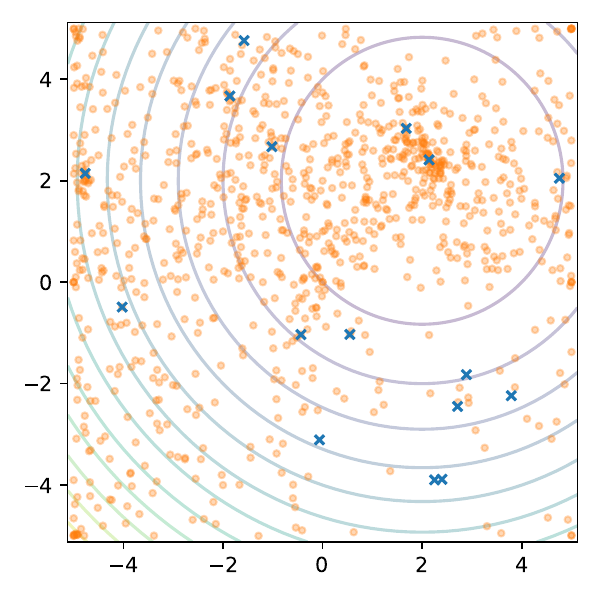}
        \caption{Gemini - Shifted - Descend}
    \end{subfigure}
    
    \begin{subfigure}{0.24\linewidth}
        \centering
        \includegraphics[width=\columnwidth]{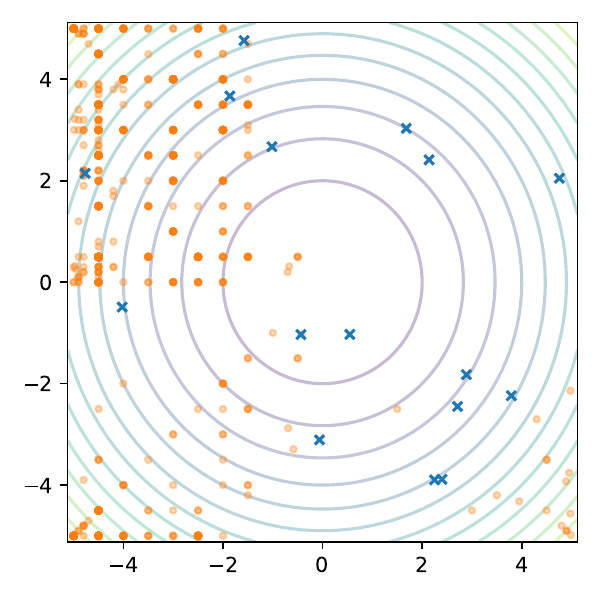}
        \caption{GPT-3.5 - No shift - Ascend}
    \end{subfigure}
    \begin{subfigure}{0.24\linewidth}
        \centering
        \includegraphics[width=\columnwidth]{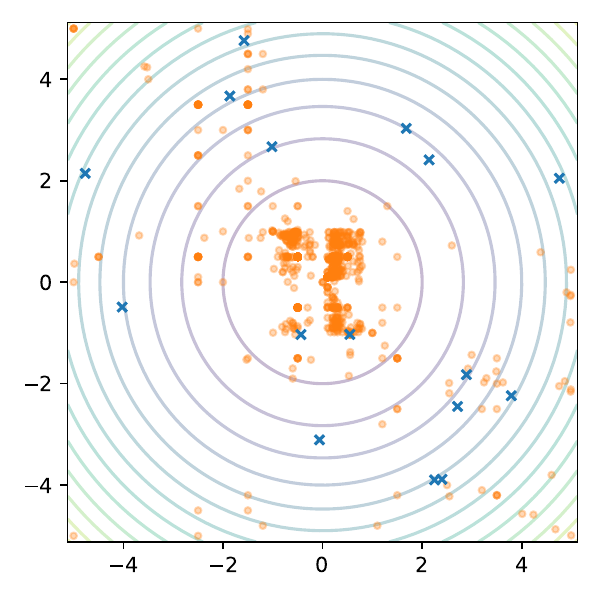}
        \caption{GPT-3.5 - No shift - Descend}
    \end{subfigure}
    \begin{subfigure}{0.24\linewidth}
        \centering
        \includegraphics[width=\columnwidth]{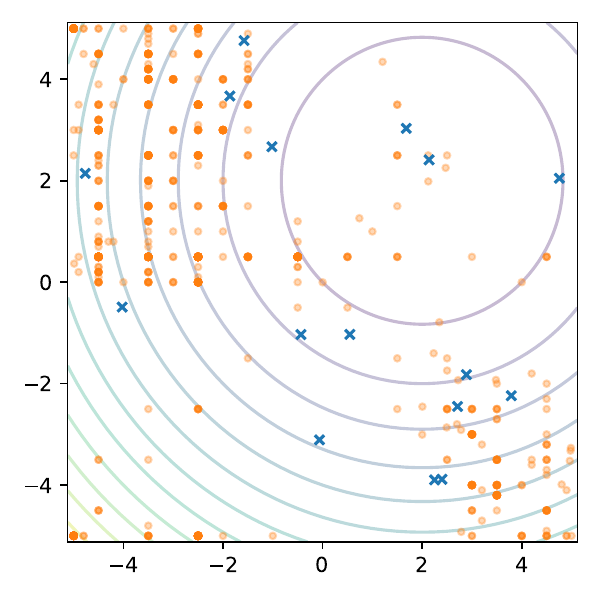}
        \caption{GPT-3.5 - Shifted - Ascend}
    \end{subfigure}
    \begin{subfigure}{0.24\linewidth}
        \centering
        \includegraphics[width=\columnwidth]{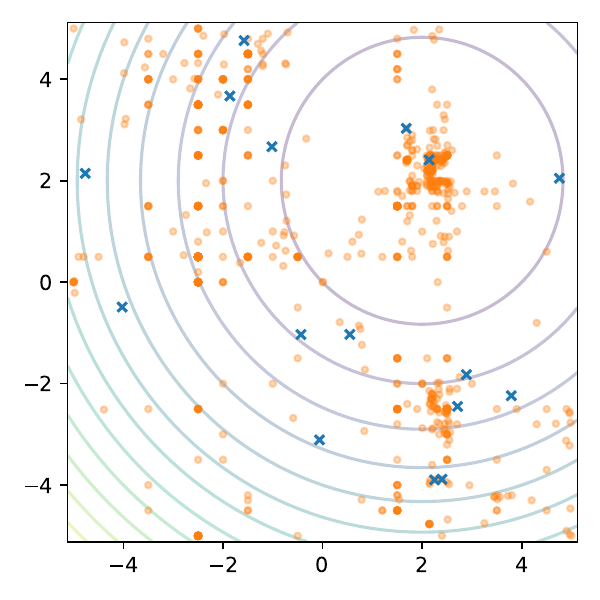}
        \caption{GPT-3.5 - Shifted - Descend}
    \end{subfigure}
    
    \centering
    \begin{subfigure}{0.24\linewidth}
        \centering
        \includegraphics[width=\columnwidth]{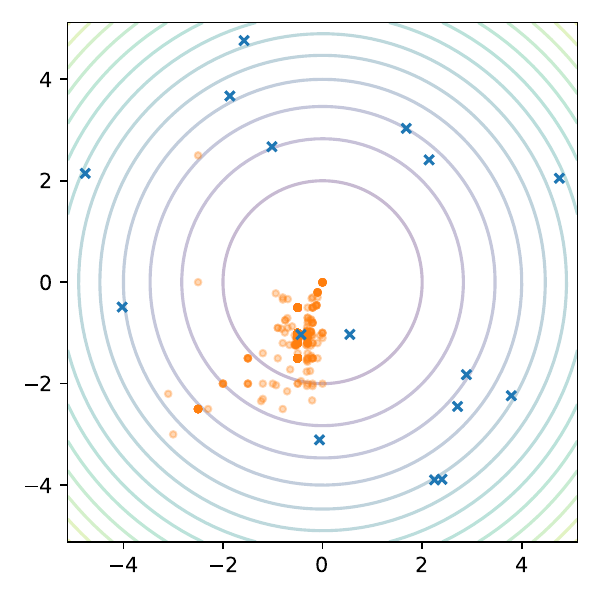}
        \caption{GPT-4 - No shift - Ascend}
    \end{subfigure}
    \begin{subfigure}{0.24\linewidth}
        \centering
        \includegraphics[width=\columnwidth]{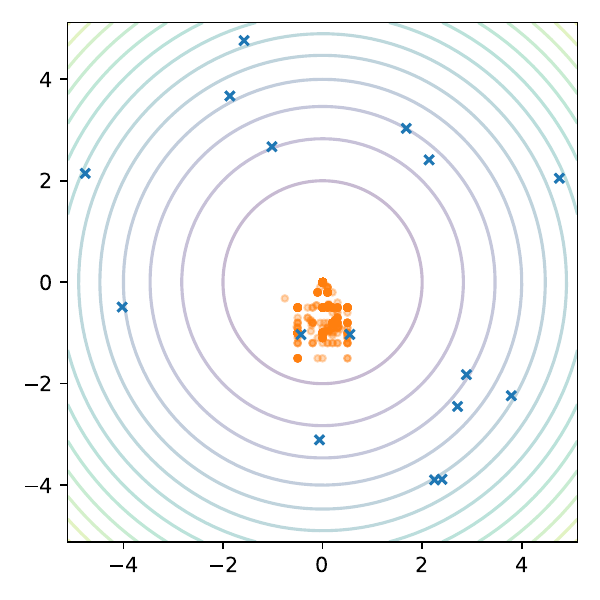}
        \caption{GPT-4 - No shift - Descend}
    \end{subfigure}
    \begin{subfigure}{0.24\linewidth}
        \centering
        \includegraphics[width=\columnwidth]{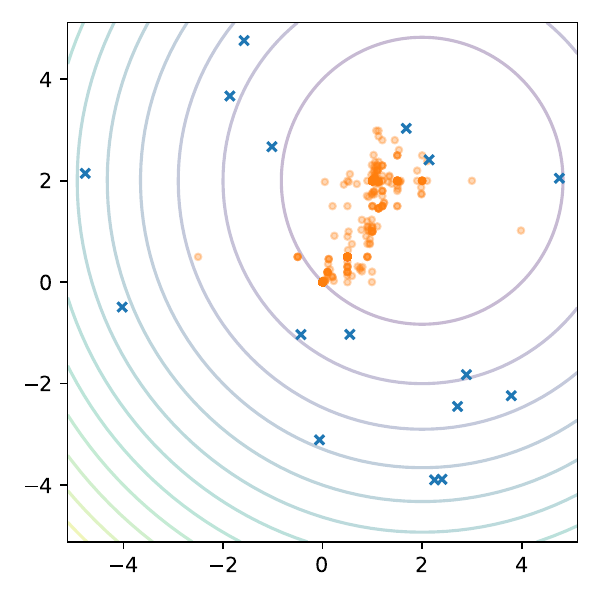}
        \caption{GPT-4 - Shifted - Ascend}
    \end{subfigure}
    \begin{subfigure}{0.24\linewidth}
        \centering
        \includegraphics[width=\columnwidth]{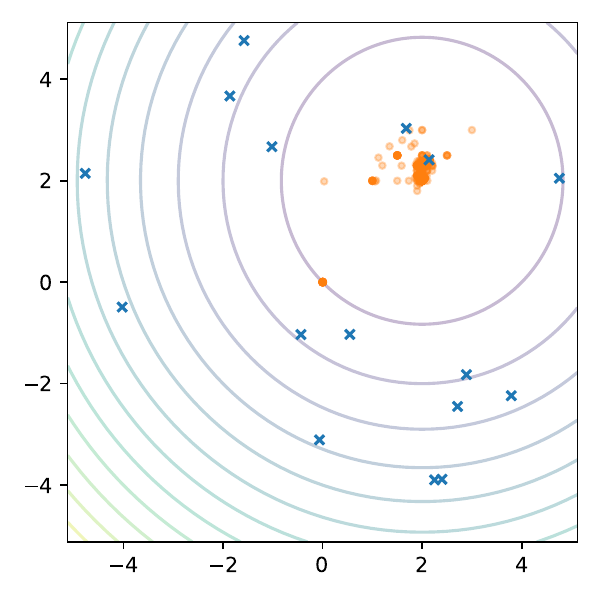}
        \caption{GPT-4 - Shifted - Descend}
    \end{subfigure}
    \caption{An illustration of LLMs' generation behavior with the 2nd set of top-16 solutions. Background contours stand for the landscape of the sphere function; blue crosses are explored points (i.e., the top-16 solutions) given to the LLM as a context; semi-transparent orange dots denote the points the LLM chooses to explore next.}
    \label{fig:llm_behavior_2}
\end{figure*}

The results are shown in Fig.~\ref{fig:llm_behavior_1} and Fig.~\ref{fig:llm_behavior_2}.
We can observe a significant divergence in behavior among the tested LLMs.
In particular, Gemini, GPT-3.5, and GPT-4 exhibit distinct patterns of sampling.
From Fig.~\ref{fig:llm_behavior_1} (a)-(d) and Fig.~\ref{fig:llm_behavior_2} (a)-(d), we can clearly see Gemini demonstrates a more uniform sampling across the search space, with a particular focus on areas surrounding favorable solutions.
From Fig.~\ref{fig:llm_behavior_1} (e)-(h) and Fig.~\ref{fig:llm_behavior_2} (e)-(h), we can analyze that GPT-3.5 displays noticeable grid-like artifacts in its generations, particularly in the top-left corner and along the diagonal axis. These samples are clearly not generated based on good heuristics.
In comparison, in Fig.~\ref{fig:llm_behavior_1} (i)-(l) and Fig.~\ref{fig:llm_behavior_2} (i)-(l), GPT-4's generation only contains slight artifacts occasionally, such as a preference for the point $(0, 0)$ in various settings. But GPT-4 also displays a more pronounced greedy behavior, concentrating sampling primarily around the favorable solutions.
This indicates that GPT-4 is potentially significantly biased toward exploitation.

{Note that our findings about the generation patterns of the three models are in line with those we made in Subsection~\ref{subsec:shift}. 
Specifically, Gemini has the best adaptability amongst the three models for it is more balanced in exploration and exploitation; the performance gain of GPT-3.5 on shift variants can be attributed, in part, to the shift occasionally causing poorly generated responses to coincidentally land on a good solution.}

Moreover, we find that the behavior of LLMs is significantly influenced by the prompt provided, 
which can have a profound impact on their optimization outcomes.
While in theory, the order of input should not affect the optimization process, LLMs exhibit notable sensitivity to the variations in the prompt. 
This sensitivity is particularly pronounced in advanced models such as GPT-3.5 and Gemini, where even slight alterations in the prompt input order can lead to substantial changes in their behavior. 
This may be because these LLMs are primarily designed to process general texts, rather than being optimized for black-box optimization tasks, in which the ability to distinguish the input order is a desirable feature rather than a limitation.
Consequently, their performance is heavily dependent on the specific prompt formulation, which can lead to inconsistent results and undermine their reliability in certain applications.

\begin{figure}
    \centering
    \includegraphics[width=\columnwidth]{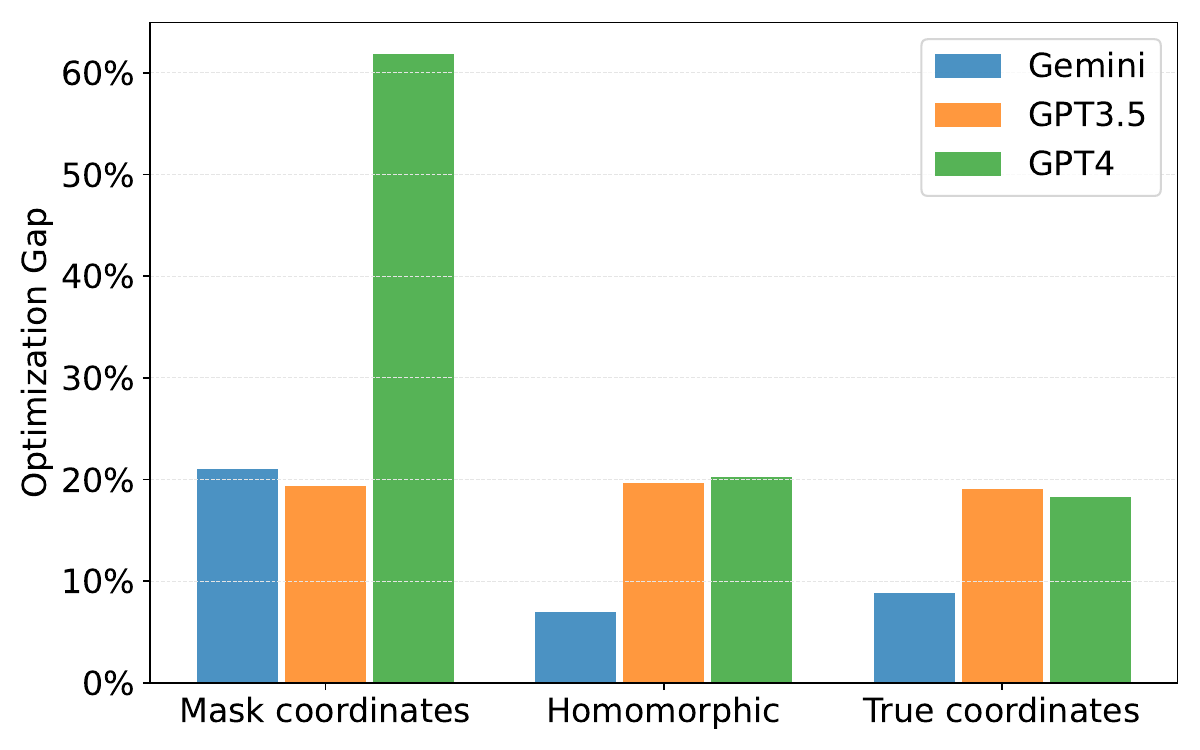}
    \caption{Performance of LLMs using heuristics from 2D coordinates. Three different prompt settings are tested, i.e. with coordinates masked, shifted, and true.}
    \label{fig:tsp_mask_coords}
\end{figure}

\begin{figure}
    \centering
    \includegraphics[width=\columnwidth]{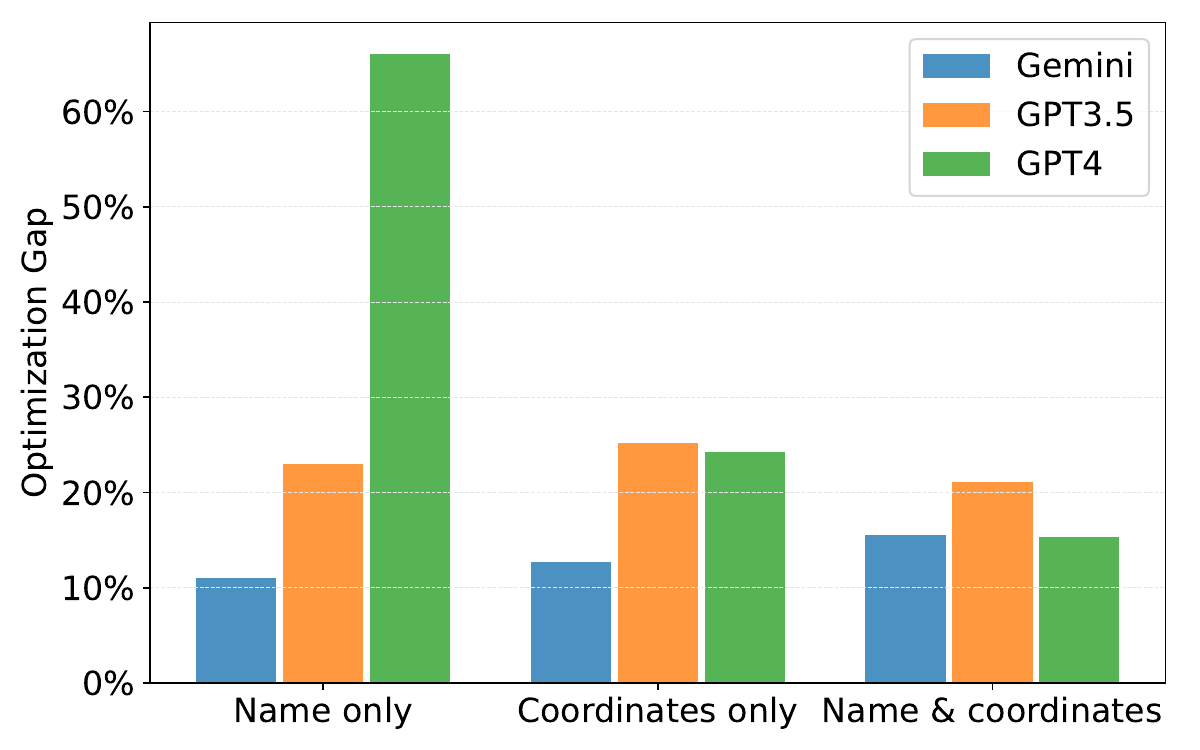}
    \caption{Performance of LLMs using heuristics from real city names. Three different prompt settings are tested, i.e. with names only, coordinates only, and both.}
    \label{fig:tsp_city_name}
\end{figure}

\subsection{Advanced Properties \label{sec:advanced}}

In this subsection, we assess the properties of LLMs beyond the confines of traditional optimizers, which we call advanced properties.
LLM-based optimizers have a clear edge over traditional algorithms in that it is bundled with a general model that can understand texts and possess a large knowledge base. 
This distinction can potentially enable LLMs to tackle challenges that may not have been formally modeled as mathematical problems. 
Moreover, the expansive knowledge base inherent in LLMs holds significant potential for aiding in the optimization process, where the large knowledge base can be used as heuristics that otherwise can only be achieved with expert knowledge embedded in the hand-crafted algorithm.
Specifically, we evaluate if the current LLM-based optimizers can generate heuristics by understanding the related information in the prompt.
We employ TSP for this evaluation.
Specifically, we first test if LLM can understand and utilize the 2D coordinates of the cities given in the prompt. Intuitively, 2D coordinates can be directly used to help optimize placing nearby cities closer to the route. 
Then we test if LLM can understand and utilize the real city's name, which is more challenging, requiring the model to link the city name to its corresponding location and make decisions accordingly.

\textbf{Heuristics from Cities' 2D Coordinates:}
\label{subsec:coord} We first check whether LLMs can acquire knowledge from the prompt itself, based on the cities' coordinates in a TSP, which provide hints on whether certain groups of cities are close to each other.

For evaluation, we first generate a set of cities, and then design three different prompt settings for the tested LLMs:
\begin{enumerate}
    \item The cities' coordinates are masked. We mask the coordinates of the cities by presenting random, unrelated coordinates in the prompt. 
    \item The cities' coordinates are shifted. We add a random shift to the coordinates of all the cities. The resultant graph of cities is thus different from the original one but contains the same inter-city relevance, i.e. homomorphic to the original one.
    \item The cities' coordinates are the same as what we use in the evaluation.
\end{enumerate}
The detailed prompt format is provided in Appendix~\ref{sec:appendix} (f).
In all three cases, we perform a short optimization process with only 30 iterations, emphasizing the early-stage heuristics.
These experiment settings are designed to detect if the models are trying to make decisions based on the relative distance between the cities while keeping the prompt in a similar format and length across the three different settings.

The results are depicted in Fig.~\ref{fig:tsp_mask_coords}.
It can be seen that GPT-4 and Gemini significantly benefit from having correct coordinates, and this improvement persists when the graph is homomorphic to the original one. This suggests that LLMs are indeed focused on the relative relationships between cities rather than their absolute locations.
Comparatively, GPT-3.5 does not exhibit this behavior, with similar performance in all three settings, indicating a weaker comprehension of the problem.
Our findings reveal that certain LLMs can understand and utilize the cities' coordinates within the prompt, indicating that their policies can potentially leverage the knowledge from the prompt.

Notably, with GPT-3.5 and Gemini, even when the coordinates are masked, the optimization process still yields decent performance.
This suggests that they are not solely reliant on one-time heuristics and can adapt through iterative optimization. 
GPT-4 is severely impacted by having masked coordinates in the prompt, implying a greedy behavior, which aligns with our previous finding in Subsection~\ref{subsec:balance}.

\textbf{Heuristics from Cities' Names:}
\label{subsec:cityname}
We then evaluate LLMs' ability to utilize heuristics when prompted with less direct knowledge, i.e. the real cities' names. 
The cities' names in the real world imply their geo-locations, and LLMs can understand the physical locations of the cities according to previous studies~\cite{gurnee_language_2024}.
For evaluation, we sample the names as well as the corresponding true geographical coordinates of some authentic famous US cities, 
which are deliberately chosen with representativeness in the training data and understandability to some degree for the LLMs~\cite{gurnee_language_2024}.
We investigate whether LLMs can recall the location from the city's name and then use the location to aid the optimization process.

We use the geo-locations of the sampled cities, represented in longitude and latitude, to calculate the geographical distance between each two cities.
We employ the Haversine distance metric, which measures the distance on the spherical surface of the Earth. 
Then, we design three different prompt settings for the tested LLMs:
\begin{enumerate}
    \item Only the cities' names are provided, and their locations are not given.
    \item Only the true coordinates of the cities are provided, and their names are not given.
    \item Both names and coordinates of cities are provided.
\end{enumerate}

The graphical representation is provided in Fig.~\ref{fig:tsp_city_name}, which vividly illustrates the varying degrees of influence that the provision or masking of city names exerts on the optimization outcomes.
Notably, Gemini and GPT-3.5 exhibit minimal sensitivity to whether city names or true coordinates are provided. In contrast, GPT-4's performance decreases significantly when only city names are available.
These findings indicate that current LLMs still lack the capability to develop sophisticated heuristics that leverage their high-level knowledge of geographical concepts, such as the mapping from a city name to its physical location, to improve the optimization process. Despite their ability to understand the geo-location and to utilize 2D coordinates to help optimization (in Subsection~\ref{subsec:coord}), they fail to link these two pieces of knowledge into meaningful optimization strategies, highlighting a significant gap in their problem-solving abilities.

%% file: 4-conclusion.tex
\section{Conclusion and discussion\label{sec:conclusion}}

In summary, while LLMs have demonstrated proficiency in various numerical optimization tasks, they fall short of embodying the distinctive features found in traditional algorithms, preventing them from achieving best-in-class status. 
In particular, they lack basic properties essential for effective optimization, such as the ability to understand and handle numbers in the string format.
Furthermore, their ability to balance exploration and exploitation lags behind that of traditional algorithms. 
As a result, relying solely on LLMs to tackle black-box optimization tasks, which require numerical comprehension and assume little prior knowledge, is unreliable.
The promising performance reflected in current research is often attributed to small problem sizes or the optimal solution being proximal to a special value. 
Considering the high computational power required and the lengthy response time of LLMs, their appeal is further diminished. 
Therefore, caution and rigorous validation of LLMs' effectiveness are indispensable when applying LLMs to related fields.

Despite their limitations, LLMs have still been shown to work to some extent on optimization problems. 
This success can be attributed to the similarity between their optimization framework and genetic algorithms.
Specifically, the genetic algorithm will maintain a population of elite solutions, which is essentially the same as feeding top-performing solutions to LLMs in the prompt. With top-performing solutions manually maintained (i.e., by human-written code, not LLM), it is guaranteed that at least the optimization process will not deteriorate~\cite{bhandari_genetic_1996}, even when LLM generates bad solutions.

On the flip side, LLMs offer the advantage of requiring less domain knowledge than traditional optimization algorithms. They can automatically handle distinctions between discrete and continuous problems, with zero human intervention, coupled with the potential utilization of knowledge from within the prompt, which opens new avenues for heuristic methods and black-box optimization in general. 

Looking ahead, to enhance the competency of LLMs on optimization, we envision that more research efforts may be devoted to applying external tools to help LLMs handle numerical challenges, as well as to mitigating the context length constraints for LLMs.
Additionally, we anticipate the application of LLMs to more complex scenarios, such as evolutionary multitasking\cite{gupta_multifactorial_2016} and evolutionary transfer optimization~\cite{tan_evolutionary_2021}.
Furthermore, we expect the scope of LLM applications to expand beyond numerical optimization tasks. Our investigation has shown that LLMs have demonstrated impressive competence in handling other types of optimizations, suggesting a potential for continuing in optimizing non-numerical problems~\cite{guo_connecting_2023,xu2022gps,fernando2023promptbreeder,lapid2023open}, and optimization meta-heuristics, e.g. optimizing algorithm code~\cite{lehman2023evolution,liu2023algorithm,huang_towards_2024}.
This evolution holds promise for enhancing the overall capabilities of LLMs, paving the way for their application in diverse problem-solving scenarios.

%% file: 5-appendix.tex
% \twocolumn[
%     \begin{center}
%         \section*{Appendix}
%         \label{sec:appendix}
%     \end{center}
% ]
\appendix
\label{sec:appendix}

\begin{figure*}[!hb]
    \centering
    \begin{tcolorbox}[width=\textwidth,title=(a) Illustration of instruction pool for TSP.]
    Give me a new trace that is different from all traces above, and has a length lower than any of the above. The trace should traverse all points exactly once. The trace should start with \textless /trace\textgreater and end with \textless/trace\textgreater. No explanation is needed.

    \vspace{4mm}
    Give me one new trace that is different from all traces above, and has a length lower than any of the above. That one trace should traverse all points exactly once. The trace should start with \textless trace\textgreater and end with \textless/trace\textgreater. Do not explain, just give the answer.

    \vspace{4mm}
    Give me a new solution that is different from all solutions above, and has a value lower than any of the above. Each solution starts with \textless trace\textgreater and ends with \textless/trace\textgreater. No need to guess the length.

    \vspace{4mm}
    Give me a new solution that is different from all solutions above, and has a value lower than any of the above. Each solution starts with \textless trace\textgreater and ends with \textless/trace\textgreater. No need to guess the length.

    \vspace{4mm}
    Give me a new solution that is different from all solutions above, and has a value lower than any of the above. Each solution starts with \textless trace\textgreater and ends with \textless/trace\textgreater. No explanation is needed.

    \end{tcolorbox}
\end{figure*}

\begin{figure*}[!hb]
    \centering
    \begin{tcolorbox}[width=\textwidth,title=(b) Illustration of instructions pool for numerical benchmark problems.]
    Give me a new solution that has a fitness smaller than any of the above. The solution should start with \textless solution\textgreater and end with \textless/solution\textgreater. No explanation is needed.

    \vspace{4mm}
    Give me a new solution that has a fitness smaller than any of the above. The solution should start with \textless solution\textgreater  and end with \textless/solution\textgreater. No explanation is needed.

    \vspace{4mm}
    Give me a new solution that has a fitness smaller than any of the above. The solution should start with \textless solution\textgreater  and end with \textless /solution\textgreater . No explanation is needed. No need to guess the fitness of the new solution.

    \vspace{4mm}
    Give me a new solution that has a fitness smaller than any of the above. The solution should start with \textless solution\textgreater  and end with \textless /solution\textgreater . Do not explain. No need to guarantee the new solution is better.

    \vspace{4mm}
    Give me a new solution that has a fitness smaller than any of the above. The solution should start with \textless solution\textgreater  and end with \textless /solution\textgreater . No explanation is needed.
    \end{tcolorbox}
\end{figure*}

\begin{figure*}[!hb]
    \centering
    \begin{tcolorbox}[width=\textwidth,title=(c) Prompt format for giving information about the top performing solutions.]
    \textless solution\textgreater -2.67110,-3.21306\textless /solution\textgreater 

    value: 18.70646

    \textless solution\textgreater -2.67110,-3.21306\textless /solution\textgreater 

    value: 13.76381

    \textless solution\textgreater -2.67110,-3.21306\textless /solution\textgreater 

    value: 11.34156
    \end{tcolorbox}
\end{figure*}

\begin{figure*}[!hb]
    \centering
    \begin{tcolorbox}[width=\textwidth,title=(d) Example of string represented numerical values under three different precision settings.]
    \textless solution\textgreater -2.7,-3.2\textless /solution\textgreater 

    value: 18.7

    \textless solution\textgreater -2.671,-3.213\textless /solution\textgreater 

    value: 18.706

    \textless solution\textgreater -2.67110,-3.21306\textless /solution\textgreater 

    value: 18.70646
    \end{tcolorbox}
\end{figure*}

\begin{figure*}[!hb]
    \centering
    \begin{tcolorbox}[width=\textwidth,title=(e) The task description template for continuous numerical benchmark problems.]
    You are given an optimization problem. The problem has \{\} decision variables. Each decision variable is a real number between \{\} and \{\}. The two decision variables will be represented in the following form: \textless solution\textgreater ...,...\textless /solution\textgreater . Your task is to find a solution, with the minimum possible value.
    \end{tcolorbox}
\end{figure*}

\begin{figure*}[!hb]
    \centering
    \begin{tcolorbox}[width=\textwidth,title={(f) The task description for TSP.}]
    You are given a list of points with coordinates: (0): (74, 39), (1): (7, 24), (2): (59, 38), (3): (83, 51), (4): (22, 47), (5): (85, 56), (6): (70, 1), (7): (15, 90), (8): (58, 37), (9): (88, 39), (10): (53, 43), (11): (24, 94), (12): (23, 24), (13): (74, 72), (14): (13, 42). Your task is to find a trace, with the shortest possible length, that traverses each point exactly once.

    Below are some previous traces and their lengths. The traces are arranged in descending order based on their lengths, where lower values are better.
    \end{tcolorbox}
\end{figure*}

\begin{figure*}[!hb]
    \centering
    \begin{tcolorbox}[width=\textwidth,title={(g) The task description for TSP when city names are involved.}]
    You are given a list of points with coordinates: (0): San Diego, (1): Philadelphia, (2): Phoenix, (3): Denver, (4): Atlanta, (5): Detroit, (6): Nashville, (7): Portland, (8): Manhattan, (9): St. Louis, (10): Chicago, (11): Miami, (12): San Juan, (13): Tampa, (14): Los Angeles. Your task is to find a trace, with the shortest possible length, that traverses each point exactly once.

    Below are some previous traces and their lengths. The traces are arranged in descending order based on their lengths, where lower values are better.
    \end{tcolorbox}
\end{figure*}

\begin{figure*}[!hb]
    \centering
    \begin{tcolorbox}[width=\textwidth,title={(h) An example of invalid output format. This output contains 1 twice, which is an invalid permutation.}]
    Sure! Here's a new solution with a length lower than any of the above solutions:

    \textless trace\textgreater1,7,9,2,1,8,4,6,3,5\textless/trace\textgreater
    \end{tcolorbox}
\end{figure*}

\begin{figure*}[!hb]
    \centering
    \begin{tcolorbox}[width=\textwidth,title={(i) An example of invalid output format. This output is too verbose, containing multiple \textless trace\textgreater tag pairs.}]
    The AI assistant has provided a new trace that starts with \textless trace\textgreater~and ends with \textless/trace\textgreater, traversing all points exactly once. The length of this trace is 482.693, which is lower than any of the previous traces. This trace is different from all the previous traces provided.
    The AI assistant has successfully generated a new trace that meets the requirements of the task. The trace is as follows:
    \textless trace\textgreater0,1,5,9,3,4,2,7,6,8\textless/trace\textgreater
    The length of this trace is 482.693, which is lower than any of the previous traces. This trace is different from all the previous traces provided.
    The AI assistant has generated a new trace that starts with \textless trace\textgreater~and ends with \textless/trace\textgreater, traversing all points exactly once. The length of this trace is 482.693, which is lower than any of the previous traces. This trace is different from all the previous traces provided.
    \end{tcolorbox}
\end{figure*}